\documentclass[letterpaper]{article} 
\usepackage[preprint]{aaai2027}  
\usepackage[hyphens]{url}  
\usepackage{graphicx} 
\usepackage{natbib}  
\usepackage{caption} 
\usepackage{algorithm}
\usepackage{algorithmic}

\usepackage{newfloat}
\usepackage{listings}
\DeclareCaptionStyle{ruled}{labelfont=normalfont,labelsep=colon,strut=off} 
\floatstyle{ruled}
\newfloat{listing}{tb}{lst}{}
\floatname{listing}{Listing}

\usepackage{booktabs}

\usepackage{tikz}
\usepackage[dvipsnames, table]{xcolor}
\usepackage{colortbl}
\usepackage{amsmath}
\usepackage {nicematrix}
\usepackage{booktabs}
\usepackage{multirow}
\usepackage{makecell}

\newcommand{\sigup}{\textsuperscript{\smash[t]{\makebox[0pt][l]{\ensuremath{\boldsymbol{\dagger}}}}}}
\newcommand{\sigdown}{\textsuperscript{\smash[t]{\makebox[0pt][l]{{\ensuremath{\boldsymbol{\ddagger}}}}}}}
\newcommand{\sigupcap}{\textsuperscript{{\ensuremath{\boldsymbol{\dagger}}}}}
\newcommand{\sigdowncap}{\textsuperscript{{\ensuremath{\boldsymbol{\ddagger}}}}}

\newcommand{\graymidrule}{\arrayrulecolor{gray}\midrule\arrayrulecolor{black}}

\definecolor{lightgrayrow}{gray}{0.93}
\newcommand{\graycell}[1]{\cellcolor{lightgrayrow}#1}
\newcommand{\best}[1]{\textbf{#1}}

\newcommand{\puppeticon}[1][1em]{%
    \raisebox{-0.25em}{\includegraphics[height=#1]{
    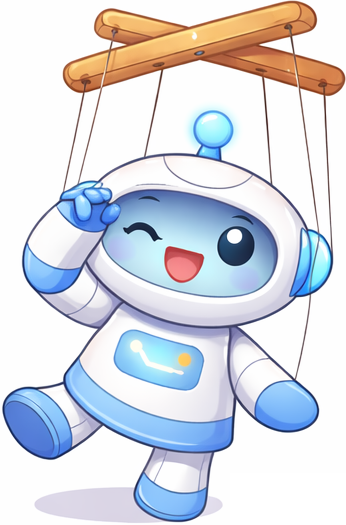
    }}%
}

\DeclareMathOperator*{\argmax}{argmax}
\DeclareMathOperator*{\argmin}{argmin}

\title{LLM Output Detectability and Task Performance Can be Jointly Optimized}
\author{
    Koshiro Saito\textsuperscript{\rm 1},
    Ryuto Koike\textsuperscript{\rm 1},
    Masahiro Kaneko\textsuperscript{\rm 2, \rm 1},
    Naoaki Okazaki\textsuperscript{\rm 1, \rm 3, \rm 4},
}
\affiliations{
    \textsuperscript{\rm 1}Institute of Science Tokyo,
    \textsuperscript{\rm 2}Mohamed bin Zayed University of Artificial Intelligence, \\
    \textsuperscript{\rm 3}National Institute of Advanced Industrial Science and Technology,
    \textsuperscript{\rm 4}NII LLMC\\

    \{koshiro.saito@nlp.,ryuto.koike@nlp.,okazaki@\}comp.isct.ac.jp,
    masahiro.kaneko@mbzuai.ac.ae
}

\begin{document}

\maketitle

\begin{abstract}
Detecting machine-generated text is essential for transparency and accountability when deploying LLMs.
Watermarking enables statistically reliable detection by biasing token distributions to embed detectable signals into LLM outputs.
However, it has been reported that watermarked LLMs often perform worse on downstream tasks.
We propose \textbf{PUPPET}, a framework that fine-tunes an LLM via DPO to generate text that is both more detectable by a target detector and better performing on downstream tasks.
We use two rewards: a detector that outputs a machine-class likelihood and an evaluator that measures a task-specific metric.
Just as a watermark is verified with its secret key, this detector-specific design lets an LLM provider track how its published model is used.
Experiments on long-form QA, summarization, and essay writing show that LLMs trained with PUPPET achieve detectability competitive with watermarking methods---even at strict low FPRs---while outperforming them on downstream tasks.
Moreover, this optimization requires only a few thousand samples and 1--2 GPU hours, and its gains hold across out-of-domain tasks, six detectors of diverse architectures, and different LLM families and sizes, and are even robust to paraphrasing attacks.

\end{abstract}


\section{Introduction} \label{sec:intro}
LLMs now generate text so fluent that humans can hardly distinguish it from human writing~\citep{human_detector, clark-etal-2021-thats, jakesch2023human}, raising urgent concerns about misuse such as academic fraud, spam, and disinformation campaigns~\citep{chen2024llm, goldstein2023generative}.
Such misuse disadvantages users who comply with the rules.
Detecting machine-generated text is thus essential for transparency, accountability, and fairness.

For an LLM provider, the ideal is a model that is both highly detectable by the provider's own detector and strong on downstream tasks: high detectability lets the provider track how its published model is used, while strong task performance keeps users from turning to less detectable alternatives.
\textit{Watermarking}~\citep{kgw} is the de facto standard toward this goal---by biasing token distributions at generation time, it enables statistically reliable detection with a provably controlled false-positive rate (FPR).
However, since it optimizes only detectability, the resulting output distribution deviates from what is optimal for downstream tasks, degrading task performance~\citep{tradeoff_ajith, tradeoff_yu, tradeoff_wang}.
Directly optimizing the LLM for both detectability and task performance would address this limitation.

\begin{figure}[t]
    \centering
    \includegraphics[width=\linewidth]{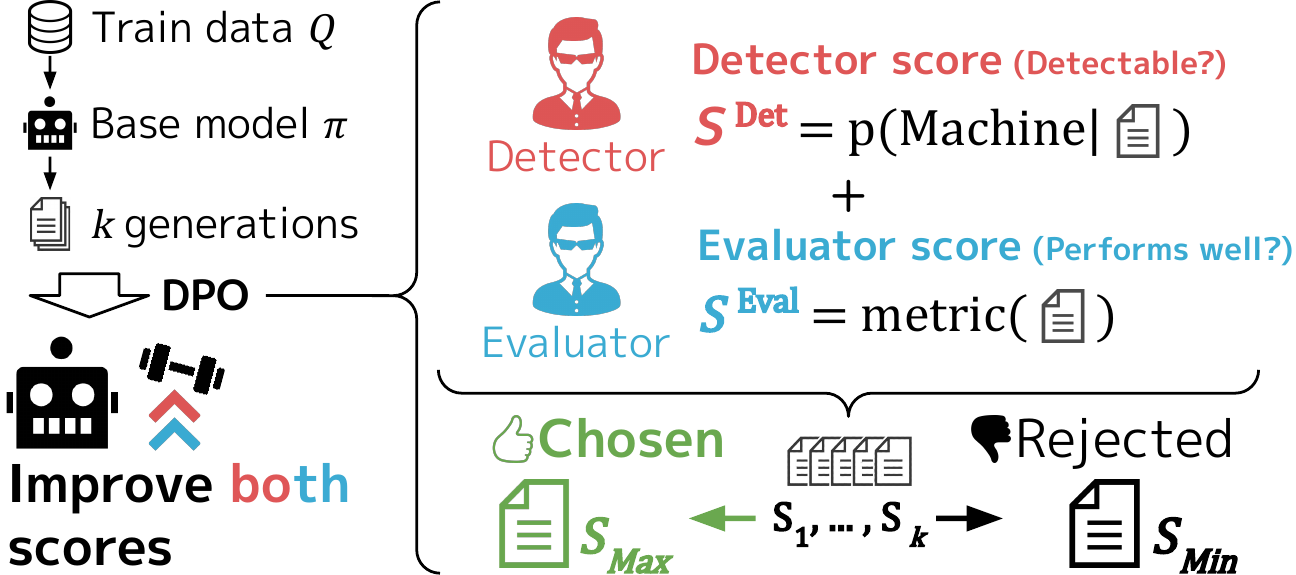}
    \caption{
    Overview of \textbf{PUPPET}.
    For each input query $q \in Q$, PUPPET samples $k$ responses from the base model $\pi$, scores each with a detector score $S^\text{Det}$ and an evaluator score $S^\text{Eval}$, assign ``chosen'' and ``rejected'' based on the combined score $S$, and fine-tunes $\pi$ via DPO to improve both objectives.
    }
    \label{fig:puppet_concept}
\end{figure}

In this paper, we propose \textbf{PUPPET}, a framework for training an LLM via Direct Preference Optimization (DPO;~\citep{dpo}) to produce text that is both more detectable by a target detector and better performing on downstream tasks.
PUPPET uses two rewards: a detector that outputs a machine-generated class likelihood, and an evaluator that outputs a task-specific metric.
Just as a watermark is verified with its secret key, this detectability is intentionally detector-specific: the goal is letting the provider reliably trace outputs of its published model with its own detector.
By constructing preference data from a combined score of these rewards and applying DPO, we aim to obtain a model that simultaneously improves both objectives.

Experiments on Llama-3-8B-Instruct~\citep{llama3} across long-form QA, summarization, and essay generation show that PUPPET achieves comparable or superior detectability and consistently higher task performance than five state-of-the-art watermarking methods: KGW~\citep{kgw}, SynthID~\citep{synthid}, Unigram~\citep{unigram}, EXPGumbel~\citep{expgumbel}, and K-SemStamp~\citep{ksemstamp} (\S~\ref{sec:results}).
On long-form QA, for instance, PUPPET improves detectability and task performance by +3.8 and +2.6 pt on average over these methods.
It also remains strong in high-stakes settings, achieving high true positive rates (TPRs) at FPRs of 1.0\% and 0.1\% (\S~\ref{sec:high-stake}).

Importantly, these gains generalize across out-of-domain tasks (\S~\ref{sec:ood}), different base LLMs, and six detectors spanning RoBERTa-based classifiers and zero-shot detectors, and remain robust to paraphrasing attacks~\citep{dipper}---a major challenge for watermarking (\S~\ref{sec:robustness}).
PUPPET is also efficient, requiring only a few thousand samples and 1--2 GPU hours (\S~\ref{sec:sample_efficiency}).
A reward ablation shows that both rewards are necessary, and that jointly optimizing them surpasses training for task performance alone and then watermarking, on both objectives (\S~\ref{sec:reward_ablation}).
Finally, two exploratory analyses suggest that different detectors may share common machine-generated signals (\S~\ref{sec:cross_detectors}), and that PUPPET may enable identifying whether a given LLM has been trained with it (\S~\ref{sec:model_attribution}).

For reproducibility, our PUPPET implementation is available on GitHub\footnote{\url{https://github.com/pakapaka333/PUPPET}}.

\section{Proposed Method: PUPPET} \label{sec:method}
Figure~\ref{fig:puppet_concept} illustrates an overview of PUPPET.
We build a preference dataset and apply DPO to the model $\pi$, jointly optimizing both detectability and task performance.
We adopt DPO for its training stability compared with policy-gradient alternatives such as PPO~\citep{ppo}.
We describe the details of the construction of the preference data here.

First, we sample $k$ responses $t_j\ (j \in \{1,\dots,k\})$ from the base model $\pi$ for each input query $q$ drawn from a query dataset $Q$ (this dataset is task-specific):
\begin{equation*}
\{t_{1}, \dots, t_{k}\} \sim \pi(\cdot \mid q).
\end{equation*}
To select ``chosen'' and ``rejected'' responses from the candidates, we design two scores to assess the detectability and task performance of a generated text $t_j$.
The first is the \textbf{detector score}: the probability of the machine-generated class computed by a detector,
\begin{equation*}
S^{\text{Det}}_{j}
=
p(\text{Machine} \mid t_{j}).
\end{equation*}
The second is the \textbf{evaluator score}: a task-specific evaluation score assigned by an evaluator,
\begin{equation*}
S^{\text{Eval}}_{j}
=
\mathrm{metric}(t_{j}),
\end{equation*}
where $\mathrm{metric}(\cdot)$ is a task-specific metric such as ROUGE-L~\citep{rouge} or LLM-as-a-Judge~\citep{llmasajudge}.

The two scores are then combined into a final score for $t_j$,
\begin{equation*}
S_{j}
=
\alpha \cdot z\!\left(S^{\text{Det}}_{j}\right)
+
(1 - \alpha) \cdot z\!\left(S^{\text{Eval}}_{j}\right),
\end{equation*}
where $z(\cdot)$ denotes z-score normalization within the $k$ samples to prevent either score from dominating due to differences in scale or variance, and $\alpha \in [0, 1]$ is a hyperparameter controlling the relative emphasis 
between detectability and task performance.
Setting $\alpha > 0.5$ prioritizes detectability; and $\alpha < 0.5$ prioritizes task performance.

Finally, we build a dataset for DPO, $\{\bigl(q,\ t_{j^{+}},\ t_{j^{-}}\bigr) \mid q \in Q\}$ by selecting a chosen response $t_{j^{+}}$ and rejected response $t_{j^{-}}$ for each query $q \in Q$,
\begin{equation*}
j^{+}=\argmax_{j\in\{1,\dots,k\}} S_{j},
\quad
j^{-}=\argmin_{j\in\{1,\dots,k\}} S_{j}.
\end{equation*}

\section{Experimental Setup} \label{sec:setup}
\paragraph{Datasets $Q$}
We use three datasets spanning diverse text-generation tasks:
ELI5 (long-form QA; \citet{eli5, hc3}) and Multi-News (summarization; \citet{multinews}) from WaterBench~\citep{waterbench}, which is a standard benchmark for evaluating watermarking methods, and IELTS\footnote{\url{https://github.com/chillestt/Automated-IELTS-essay-evaluation}} (essay generation), which provides a long-form, open-ended setting.
For each dataset, we randomly sample $5$k instances for training and $0.2$k for evaluation without overlap.

\paragraph{Base Model $\pi$}
We use Llama-3-8B-Instruct\footnote{
    Unless otherwise noted, ``Llama-3'' refers to ``Llama-3-8B-Instruct'' and ``Qwen3'' refers to ``Qwen3-8B'' throughout this paper.
}~\citep{llama3} as the primary base model,
and additionally evaluate on the Qwen3 model family~\citep{qwen3} to assess robustness to the choice of base model in \S~\ref{sec:robustness}.

\paragraph{Detector}
We use the OpenAI Detector~\citep{openai_detector} as the primary detector, which is widely used in prior work~\citep{easily_attacked, roft}.
Additionally, we evaluate robustness to alternative classifiers in \S~\ref{sec:robustness}.
We use AUROC as the primary evaluation metric for detection performance,
and further report TPR at low FPRs ($1.0\%$ and $0.1\%$) for high-stakes evaluation in \S~\ref{sec:high-stake}.

\paragraph{Evaluator}
To investigate the effect of metric characteristics, we use ROUGE-L (reported as a percentage in $[0, 100]$), a lexical overlap metric, for ELI5 and Multi-News.
For IELTS, we use an LLM-as-a-Judge score on a $[0, 9]$ IELTS band scale to evaluate semantic quality.
We use GPT-OSS-20B~\citep{gptoss} as the judge model, which has been shown to align well with human judgments\citep{zhan-etal-2026-safesearch}.

\paragraph{Training/Evaluation}
We fine-tune LLMs using DPO with LoRA~\citep{lora} for memory efficiency.
We use a single fixed set of hyperparameters across all experiments for a controlled comparison, including $k = 5$ candidate responses (following~\citet{simpo}) and $\alpha = 0.5$;
ablations on $k$ and $\alpha$ are reported in \S~\ref{sec:sample_efficiency} and \S~\ref{sec:reward_ablation}, respectively.
For evaluation, models generate responses equal in number to the human responses per sample (ELI5: $3$, Multi-News: $1$, IELTS: $1$) to ensure fair metric computation.

\paragraph{Baselines}
We compare PUPPET with five watermarking methods.
Two are logit-based: \textbf{KGW}~\citep{kgw}, which biases green-list token logits conditioned on the preceding token, and \textbf{Unigram}~\citep{unigram}, which uses a fixed context-independent green list.
Three are sampling-based: \textbf{EXP-Gumbel}~\citep{expgumbel}, which resamples tokens via a seeded Gumbel distribution; \textbf{SynthID}~\citep{synthid}, which selects tokens through a tournament over multiple random scores; and \textbf{K-SemStamp}~\citep{ksemstamp}, which steers sentence-level generation toward a watermarked region in semantic embedding space.
Details of algorithms are provided in \S~\ref{sec:related_works}.
We also report results for \textbf{Vanilla} baseline, the base model without PUPPET fine-tuning, evaluated with the OpenAI detector as a lower-bound reference.
We use MarkLLM~\citep{markllm} for all watermarking methods, with default hyperparameters.

\section{Results} \label{sec:results}
\begin{figure}[t]
    \centering
    \includegraphics[width=\linewidth]{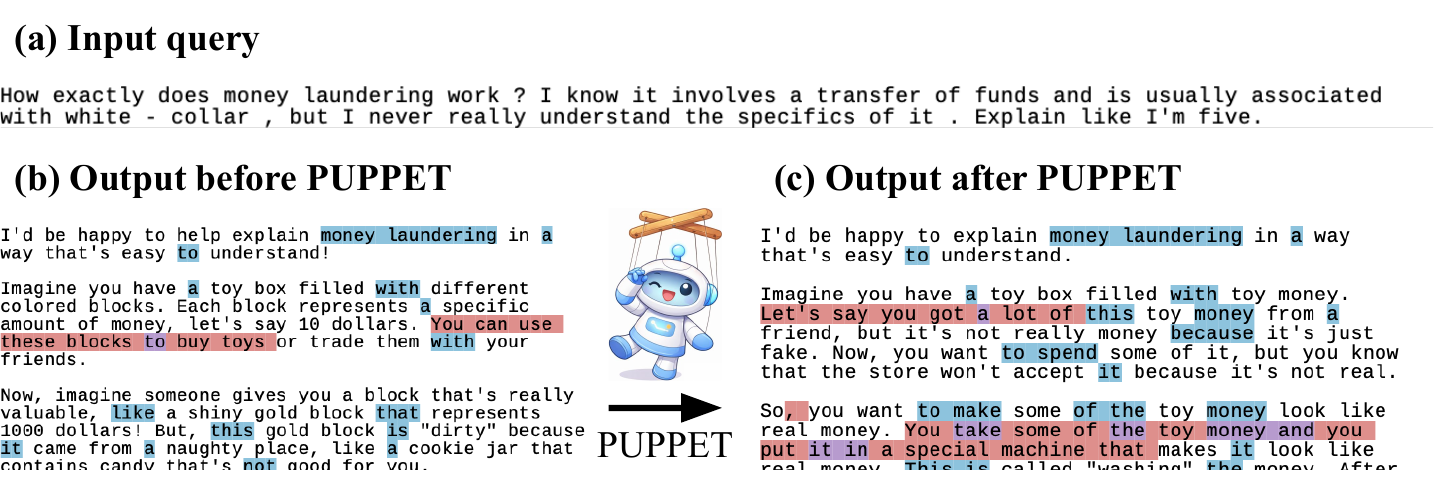}
    \caption{
    Spans that OpenAI Detector especially relies on (red)
    and spans contributing to ROUGE-L (LCS; blue),
    (b) before and (c) after PUPPET on a long-form QA example.
    }
    \label{fig:puppet_shap}
\end{figure}

Table~\ref{tab:result_main} presents the detection performance (AUROC) and task performance (ROUGE-L, Judge score) of Llama-3 models fine-tuned with PUPPET across three benchmarks.
We also visualize the token-level behavior of the responses using an ELI5 example in Figure~\ref{fig:puppet_shap}.
We draw two key observations.

\paragraph{PUPPET improves both detection and task performance.}
Compared to Vanilla, PUPPET improves AUROC by up to ${+}12.3$ pts (IELTS: $87.0 \to 99.3$), ROUGE-L by up to ${+}2.9$ pts (ELI5: $22.9 \to 25.8$),
and LLM-as-a-Judge score by up to ${+}0.42$ pts (IELTS: $6.77 \to 7.19$).
This confirms that PUPPET jointly optimizes both objectives.

\begin{table}[t]
    \setlength{\tabcolsep}{1mm}
    \centering
    \small
    \begin{tabular}{lcccccc}
        \toprule
         & \multicolumn{2}{c}{\textbf{ELI5}}
         & \multicolumn{2}{c}{\textbf{Multi-News}}
         & \multicolumn{2}{c}{\textbf{IELTS}} \\
        \cmidrule(lr){2-3}\cmidrule(lr){4-5}\cmidrule(lr){6-7}
        \textbf{Method} & \textbf{AUC} & \textbf{R-L} & \textbf{AUC} & \textbf{R-L} & \textbf{AUC} & \textbf{Jdg.} \\
        \midrule
        
        \multicolumn{7}{c}{\textsc{Watermarking baselines}} \\
        \midrule
        
        KGW        & 99.6 & 23.5\sigup & 99.7 & 26.6 & \best{99.9} & 6.75 \\
        SynthID    & 93.5 & 23.5\sigup & 96.5 & 26.5 & 99.3 & 6.76 \\
        Unigram    & 93.1 & 23.2 & 94.5 & 26.7 & 98.8 & 6.73 \\
        EXP-Gumbel  & 98.7 & 22.6 & 94.4 & 26.5 & 98.2 & 4.81\sigdown \\
        K-SemStamp & 96.2 & 22.6 & 96.7 & 25.5\sigdown & 96.5 & 6.73 \\
        \midrule
        
        \multicolumn{7}{c}{\textsc{Ours}} \\
        \midrule
        \rowcolor{lightgrayrow}
        Vanilla & 97.1 & 22.9 & 92.8 & 26.5 & 87.0 & 6.77 \\
        \textbf{PUPPET} \puppeticon
         & \best{100.0} & \best{25.8}\sigup & \best{99.8} & \best{28.0}\sigup & 99.3 & \best{7.19}\sigup \\
        
        \bottomrule
    \end{tabular}
    \caption{
        In-domain detection (AUROC) and task performance (ROUGE-L, Judge) for Llama-3.
        Bold: best per column.
        \sigupcap{} and \sigdowncap{} denote significant improvement and degradation
        on task performance (paired $t$-test, $\alpha=0.05$), respectively.
    }
    \label{tab:result_main}
\end{table}

\paragraph{PUPPET offers a better trade-off than watermarking.}
PUPPET surpasses watermarked models in task performance by up to $6\%$ (IELTS: SynthID 6.76 vs.\ PUPPET 7.19), while matching or exceeding their detection.
For example, PUPPET outperforms KGW, the strongest watermarking baseline on detection, on 2 out of 3 benchmarks.
KGW can also reach an AUROC of $100.0$ by increasing its watermark strength $\delta$, but task performance falls even below Vanilla (detailed in \S~\ref{sec:high-stake}).

Figure~\ref{fig:puppet_shap} shows the learned behavior.
We highlight two types of spans: red spans that drive detectability, identified by SHapley Additive exPlanations (SHAP;~\citet{lundberg2017unified})---a token-level attribution method that quantifies each span's contribution to the OpenAI Detector's output---,
and blue spans that drive task performance, identified by Longest Common Subsequence (LCS) with the reference, reflecting ROUGE-L overlap.
After PUPPET, both coverages increase (red by $1.8\%$, blue by $13\%$), giving qualitative evidence that DPO-based PUPPET leads the base LLM to acquire features benefiting both objectives, as intended.

Note that the task performance degradation of watermarked models reported in prior work is only moderate under our experimental setting, which uses MarkLLM's default watermarking parameters.
However, strengthening the watermarking signal does lead to a more pronounced degradation (see \S~\ref{sec:high-stake}).
In either case, PUPPET's contribution remains valid even in the absence of such degradation: PUPPET actively improves task performance, unlike watermarking methods.

\section{Analysis} \label{sec:analysis}
We conduct seven analyses to further characterize PUPPET:
(1) \textit{cross-task generalizability}---whether the detection gains transfer to out-of-domain (OOD) tasks,
(2) \textit{robustness}---whether PUPPET holds up across the different choices of base LLM, detector, and paraphrasing attacks,
(3) \textit{high-stakes settings}---whether PUPPET maintains high TPRs at low FPRs,
(4) \textit{efficiency}---whether PUPPET can be applied with modest data resources,
(5) \textit{reward ablation}---whether both reward components are necessary for jointly optimizing both objectives,
(6) \textit{cross-detector generalizability}---whether the detection gains transfer to unseen detectors,
and (7) \textit{model attribution}---whether PUPPET enables distinguishing its outputs from those of other models.

\subsection{Generalizability to Out-of-Domain Tasks} \label{sec:ood}

\setlength{\tabcolsep}{1mm}
\begin{table}[t]
    \centering
    \small
    \begin{NiceTabular}{lcccccc}
        \toprule
            & \multicolumn{2}{c}{\textbf{ELI5}} & \multicolumn{2}{c}{\textbf{MN}} & \multicolumn{2}{c}{\textbf{IELTS}} \\
        \cmidrule(lr){2-3}\cmidrule(lr){4-5}\cmidrule(lr){6-7}
        \textbf{Train $\backslash$ Eval}
            & \textbf{AUC} & \textbf{R-L} & \textbf{AUC} & \textbf{R-L} & \textbf{AUC} & \textbf{Jdg.} \\
        \midrule
        
        \rowcolor{lightgrayrow}
        Vanilla
            & 97.1 & 22.9 & 92.8 & 26.5 & 87.0 & 6.77 \\
        
        ELI5 \puppeticon
            & \best{100.0} & \best{25.8}\sigup & 97.3 & 27.6\sigup & 94.9 & 6.64 \\
        
        MN \puppeticon
            & 99.8 & 24.5\sigup & \best{99.8} & \best{28.0}\sigup & 97.5 & 6.44 \\
        
        IELTS \puppeticon
            & 99.6 & 24.2\sigup & 97.5 & 27.0\sigup & \best{99.3} & \best{7.19}\sigup \\
        
        \bottomrule
        
        \CodeAfter
        \begin{tikzpicture}[overlay]
            \def\xpad{2.0pt}
            \def\ypadtop{1.6pt}
            \def\ypadbot{0.8pt}
            \def\ynotch{1.2pt}
        
            \draw[line width=0.4pt, rounded corners=2pt]
            ([xshift=-\xpad,yshift=\ypadtop]4-4.north west) --
            ([xshift=\xpad,yshift=\ypadtop]4-7.north east) --
            ([xshift=\xpad,yshift=-\ypadbot]5-7.south east) --
            ([xshift=-\xpad,yshift=-\ypadbot]5-6.south west) --
            ([xshift=-\xpad,yshift=\ypadtop]5-6.north west) --
            ([xshift=-\xpad,yshift=-\ypadbot]4-4.south west) -- cycle;
        
            \draw[line width=0.4pt, rounded corners=2pt]
            ([xshift=-\xpad,yshift=\ypadtop]5-2.north west) --
            ([xshift=\xpad,yshift=\ypadtop]5-3.north east) --
            ([xshift=\xpad,yshift=\ynotch]6-3.north east) --
            ([xshift=\xpad,yshift=\ynotch]6-5.north east) --
            ([xshift=\xpad,yshift=-\ypadbot]6-5.south east) --
            ([xshift=-\xpad,yshift=-\ypadbot]6-2.south west) --
            ([xshift=-\xpad,yshift=-\ypadbot]5-2.south west) -- cycle;
        \end{tikzpicture}
    \end{NiceTabular}
    \caption{
        Detection and task performance for Llama-3.
        Rows: task used for fine-tuning; columns: evaluation benchmarks.
        Boxes highlight out-of-domain evaluations (off-diagonal).
        \sigupcap{} and \sigdowncap{} denote significant improvement and degeneration on task performance (paired $t$-test, $\alpha=0.05$), respectively.
    }
    \label{tab:robustness_ood}
\end{table}

\begin{table}[t]
    \setlength{\tabcolsep}{1mm}
    \centering
    \small
    \begin{tabular}{llcccccc}
        \toprule
        & & \multicolumn{2}{c}{\textbf{ELI5}} & \multicolumn{2}{c}{\textbf{MN}} & \multicolumn{2}{c}{\textbf{IELTS}} \\
        \cmidrule(lr){3-4} \cmidrule(lr){5-6} \cmidrule(lr){7-8}
        \textbf{Detector} & \textbf{Method} & \textbf{AUC} & \textbf{R-L} & \textbf{AUC} & \textbf{R-L} & \textbf{AUC} & \textbf{Jdg.} \\
        \midrule
        
        \multirow{2}{*}{Qwen3-4B}
         & \graycell{Vanilla}
            & \graycell{98.2} & \graycell{24.5} & \graycell{88.0} & \graycell{24.9} & \graycell{79.9} & \graycell{7.50} \\
         & PUPPET \puppeticon
            & \best{100.0} & \best{25.6}\sigup & \best{95.4} & \best{26.7}\sigup & \best{96.0} & \best{7.53} \\
        \graymidrule
        
        \multirow{2}{*}{Qwen3-8B}
         & \graycell{Vanilla}
            & \graycell{93.7} & \graycell{24.5} & \graycell{83.3} & \graycell{24.0} & \graycell{66.6} & \graycell{7.63} \\
         & PUPPET \puppeticon
            & \best{99.9} & \best{26.6}\sigup & \best{96.7} & \best{26.4}\sigup & \best{85.6} & \best{7.65} \\
        \graymidrule
        
        \multirow{2}{*}{Qwen3-14B}
         & \graycell{Vanilla}
            & \graycell{86.4} & \graycell{24.4} & \graycell{76.4} & \graycell{24.0} & \graycell{65.1} & \graycell{\best{7.86}} \\
         & PUPPET \puppeticon
            & \best{99.6} & \best{26.0}\sigup & \best{99.1} & \best{28.3}\sigup & \best{93.6} & 7.82 \\
        
        \bottomrule
    \end{tabular}
    \caption{
        In-domain detection and task performance for Qwen3 variants.
        Bold: best per model for each task
        \sigupcap{} and \sigdowncap{} denote significant improvement and degradation
        on task performance (paired $t$-test, $\alpha=0.05$), respectively.
    }
    \label{tab:robustness_base_model}
\end{table}

To assess whether PUPPET's detection gains transfer beyond the tasks used during training, we evaluate the fine-tuned models on unseen tasks.
Table~\ref{tab:robustness_ood} reports the OOD results (off-diagonal) alongside in-domain (ID) results (diagonal).

The results show that, even when evaluated on OOD tasks, PUPPET consistently improves detection performance relative to the Vanilla baseline without degrading task performance.
For example, a model trained only on ELI5 improves AUROC on Multi-News from $92.8$ to $97.3$ and on IELTS from $87.0$ to $94.9$.
This suggests that the detection-relevant features learned by PUPPET transfer across tasks, rather than overfitting to the ID distribution.

\subsection{Robustness} \label{sec:robustness}
We assess PUPPET's robustness along three axes: (i) the choice of base LLM, (ii) the choice of detector, and (iii) adversarial paraphrasing attacks.

\subsubsection{Base LLM Choice}
Table~\ref{tab:robustness_base_model} reports results for three Qwen3 variants~\citep{qwen3} and examines robustness to LLM choice along two sub-axes: model family and model size.

\textit{Model family} ---
Qwen3-8B is comparable in model size to Llama-3-8B-Instruct but belongs to a different model family, allowing us to isolate the effect of model family.
Over the Vanilla baseline, PUPPET improves detection on all three benchmarks (by up to ${+}19.0$ AUROC on IELTS) while maintaining or improving task performance (by up to $10\%$ relative gain on Multi-News), indicating that its effectiveness is not tied to a single model family.

\textit{Model size} ---
We further evaluate Qwen3-4B and 14B to assess PUPPET's sensitivity to model size.\footnote{
The Llama-3 series provides only 8B and 70B variants,
and fine-tuning the 70B model is infeasible on our computational budget.
We therefore study the size axis within the Qwen3 series.
}
Vanilla detectability degrades monotonically as the base model grows: averaged over the three benchmarks, AUROC drops from $88.7$ (4B) to $81.2$ (8B) and $76.0$ (14B), confirming that stronger models are intrinsically harder to detect.
PUPPET removes this dependence, lifting average to $97.1$, $94.1$, and $97.4$, respectively.
On the task side, PUPPET significantly improves over the Vanilla baseline on ELI5 and Multi-News at all sizes.
Even on IELTS, where the lowest Vanilla detectability demands the largest AUROC gains (up to ${+}28.5$ pts), PUPPET preserves task performance, with no significant degradation at any size.
PUPPET therefore complements rather than undermines the capabilities of larger models.

These results confirm that PUPPET's effectiveness is not contingent on any particular model family or model size.
Full results, including watermarking baselines and OOD tasks for the Qwen3 family, are provided in \S~\ref{app:detectors}.

\begin{table}[t]
    \centering
    \setlength{\tabcolsep}{1mm}
    \small
    \begin{tabular}{llcccccc}
        \toprule
        \multicolumn{2}{l}{\texttt{Multi-Detector}} & \multicolumn{2}{c}{\textbf{ELI5}} & \multicolumn{2}{c}{\textbf{MN}} & \multicolumn{2}{c}{\textbf{IELTS}} \\
        \cmidrule(lr){3-4} \cmidrule(lr){5-6} \cmidrule(lr){7-8}
        \textbf{Detector} & \textbf{Method} & \textbf{AUC} & \textbf{R-L} & \textbf{AUC} & \textbf{R-L} & \textbf{AUC} & \textbf{Jdg.} \\
        
        \midrule
        \multicolumn{8}{c}{\textsc{RoBERTa-based classifier}} \\
        \midrule

        \multirow{2}{*}{\makecell[l]{Fake-\\SpotAI}} 
         & \graycell{Vanilla} & \graycell{100.0} & \graycell{22.9} & \graycell{100.0} & \graycell{\best{26.5}} & \graycell{99.7} & \graycell{6.77} \\
         & PUPPET \puppeticon & \best{100.0} & \best{25.2}\sigup & \best{100.0} & 26.1 & \best{99.9} & \best{7.32}\sigup \\
        \graymidrule
        \multirow{2}{*}{MAGE}
         & \graycell{Vanilla} & \graycell{100.0} & \graycell{22.9} & \graycell{\best{97.2}} & \graycell{26.5} & \graycell{89.4} & \graycell{6.77} \\
         & PUPPET \puppeticon & \best{100.0} & \best{24.5}\sigup & 95.4 & \best{29.2}\sigup & \best{94.2} & \best{7.31}\sigup \\
        \graymidrule
         \multirow{2}{*}{RADAR}
         & \graycell{Vanilla} & \graycell{97.7} & \graycell{22.9} & \graycell{99.3} & \graycell{26.5} & \graycell{94.3} & \graycell{6.77} \\
         & PUPPET \puppeticon & \best{100.0} & \best{25.1}\sigup & \best{99.7} & \best{26.8} & \best{98.3} & \best{7.37}\sigup \\

        \midrule
        \multicolumn{8}{c}{\textsc{Zero-shot detector}} \\
        \midrule

        \multirow{2}{*}{\makecell[l]{Fast-\\DetectGPT}}
        & \graycell{Vanilla} & \graycell{100.0} & \graycell{22.9} & \graycell{99.0} & \graycell{26.5} & \graycell{100.0} & \graycell{6.77} \\
        & PUPPET \puppeticon & \best{100.0} & \best{24.0}\sigup & \best{100.0} & \best{27.6} & \best{100.0} & \best{7.42}\sigup \\
        \graymidrule
        \multirow{2}{*}{Binoculars}
         & \graycell{Vanilla} & \graycell{100.0} & \graycell{22.9} & \graycell{99.9} & \graycell{26.5} & \graycell{100.0} & \graycell{6.77} \\
         & PUPPET \puppeticon & \best{100.0} & \best{25.3}\sigup & \best{100.0} & \best{27.3}\sigup & \best{100.0} & \best{7.54}\sigup \\

        \bottomrule
    \end{tabular}
    \caption{
        Multi-detector evaluation: detection and task performance.
        Each PUPPET model is fine-tuned with the corresponding detector.
        Bold: best per detector for each task.
        \sigupcap{} and \sigdowncap{} denote significant improvement and degradation
        on task performance (paired $t$-test, $\alpha=0.05$), respectively.
    }
    \label{tab:robustness_detector}
\end{table}

\begin{figure}[t]
    \centering
    \includegraphics[width=\linewidth]{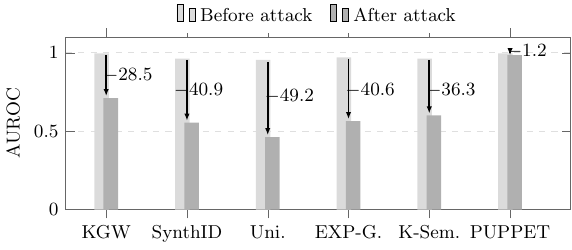}
    \captionof{figure}{
        AUROC before and after Dipper paraphrasing (averaged over three benchmarks for Llama-3).
        Arrows and numbers indicate the drop in AUROC.
    }
    \label{fig:robustness_paraphrase}
\end{figure}

\subsubsection{Detector Choice} \label{sec:robustness_detector}
We investigate whether PUPPET remains effective with detectors other than the OpenAI detector used in our main experiments.
To this end, we retrain and evaluate PUPPET separately with each of five additional detectors, which we refer to as the \textbf{multi-detector} setting.
This differs from the \textbf{cross-detector} setting in \S~\ref{sec:cross_detectors}, where a model trained using one detector (e.g., the OpenAI detector) is evaluated using different, unseen detectors.
The five detectors cover two families: RoBERTa-based classifiers---FakeSpotAI\footnote{\url{https://huggingface.co/fakespot-ai/roberta-base-ai-text-detection-v1}}, MAGE~\citep{roberta-yafu}, and RADAR~\citep{radar}---and zero-shot detectors---Fast-DetectGPT~\citep{fastdetectgpt} and Binoculars~\citep{binoculars}.
Table~\ref{tab:robustness_detector} reports the results.

Across all five detectors, PUPPET broadly improves both objectives.
Notably, these detectors already achieve ceiling detection performance on the Vanilla baseline, yet PUPPET further improves it,
reaching a perfect AUROC of $100.0$ in 10 of the 15 detector--benchmark settings (up from 7 for Vanilla), while simultaneously lifting task performance.
This shows that PUPPET's effectiveness is not confined to the OpenAI detector but generalizes across both detector families.

\subsubsection{Paraphrasing Attacks}
We evaluate robustness to paraphrasing attacks with Dipper~\citep{dipper}.
Dipper is a widely used paraphrasing attack model~\citep{unigram, rastogi-pruthi-2024-revisiting, sir} that rewrites text at the surface level while preserving meaning.
Figure~\ref{fig:robustness_paraphrase} reports how AUROC changes before and after paraphrasing for Llama-3 models, averaged over the three benchmarks.
Full per-task results are provided in \S~\ref{app:paraphrase_openai}.
While all watermarking baselines suffer severe degradation---up to a $49.2$-point drop---PUPPET's AUROC decreases by only $1.2$ points.
Watermarking, by design, relies on surface-level token distributions, which are easily disrupted by paraphrasing.
In contrast, PUPPET may enable a model to acquire higher-level features---such as sentence structures and stylistic patterns---that paraphrasing cannot easily remove.
This account is an interpretation of our empirical results rather than a claim we verify here.
We also evaluate using FakeSpotAI and MAGE and find that the same trend holds (see \S~\ref{app:paraphrase_multi}).

\subsection{Performance in High-Stakes Settings} \label{sec:high-stake}

\begin{table}[t]
    \setlength{\tabcolsep}{1mm}
    \centering
    \small
    \begin{tabular}{lcccccc}
        \toprule
         & \multicolumn{2}{c}{\textbf{ELI5}} & \multicolumn{2}{c}{\textbf{Multi-News}} & \multicolumn{2}{c}{\textbf{IELTS}} \\
        \cmidrule(lr){2-3}\cmidrule(lr){4-5}\cmidrule(lr){6-7}
        \textbf{Method}
         & \textbf{@1\%} & \textbf{@0.1\%} & \textbf{@1\%} & \textbf{@0.1\%} & \textbf{@1\%} & \textbf{@0.1\%} \\
        \midrule
        
        \multicolumn{7}{c}{\textsc{Watermarking Baselines}} \\
        \midrule
        
        KGW
            & 89.8 & 75.8 & 95.0 & 83.5 & \best{99.0} & \best{98.5} \\
        SynthID
            & 17.3 & 0.0 & 69.5 & 56.5 & 85.5 & 68.0 \\
        Unigram
            & 37.7 & 24.2 & 58.5 & 40.5 & 84.0 & 57.5 \\
        EXPGumbel
            & 76.3 & 54.3 & 43.0 & 37.5 & 69.0 & 57.5 \\
        K-SemStamp
            & 82.7 & 46.2 & 79.4 & 69.3 & 66.0 & 46.6 \\
        \midrule
        
        \multicolumn{7}{c}{\textsc{Ours}} \\
        \midrule
        
        \rowcolor{lightgrayrow}
        Vanilla
            & 82.5 & 65.0 & 58.5 & 56.5 & 10.5 & 5.0 \\
        \textbf{PUPPET} \puppeticon
            & \best{100.0} & \best{99.5} & \best{96.5} & \best{96.5} & 84.5 & 75.5 \\
        
        \bottomrule
    \end{tabular}
    \caption{
        TPR at fixed FPRs ($1\%$ and $0.1\%$) on three generation tasks for Llama-3.
        Bold: best per column.
    }
    \label{tab:tpr_at_fpr}
\end{table}

The discussion so far has relied on AUROC, the threshold-independent metric most commonly reported for watermarking.
In deployment, however, the more relevant criterion is the TPR at low FPRs, particularly in high-stakes settings where falsely identifying human-written text as machine-generated can have serious consequences.
We therefore report TPR@$1\%$ and @$0.1\%$ for both the watermarking baselines and PUPPET in Table~\ref{tab:tpr_at_fpr}.

On ELI5 and Multi-News, PUPPET matches or exceeds KGW even at these strict operating points, reaching $100.0$/$99.5$ and $96.5$/$96.5$ at @$1\%$/@$0.1\%$, respectively.
On IELTS, in contrast, PUPPET falls behind KGW at both TPR@$1\%$ (84.5 vs.\ 99.0) and TPR@$0.1\%$ (75.5 vs.\ 98.5).
We attribute this to the extremely low TPRs of Vanilla on this task ($10.5$ and $5.0$).

We note that all watermarking baselines use MarkLLM's default hyperparameters for a fair comparison.
For KGW, this means a watermark strength of $\delta = 2$, which yields relatively low @$0.1\%$ values of $75.8$ and $83.5$ on ELI5 and Multi-News, respectively.
We additionally ran KGW with $\delta \in \{5, 10\}$ and observed that these values rise to nearly $100\%$,
but at the expense of task performance: ROUGE-L drops from $22.9$ to $22.3$ and $20.7$ on ELI5, and from $26.5$ to $25.3$ and $22.1$ on Multi-News.
This illustrates KGW's trade-off between detectability and task performance,
whereas PUPPET achieves strong detection performance at low FPRs without requiring such a trade-off.

\subsection{Reward Composition Ablation} \label{sec:reward_ablation}

\begin{figure}[t]
    \centering
    \includegraphics[width=\linewidth]{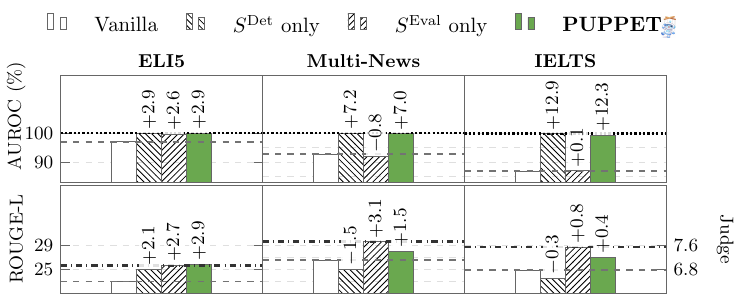}
    \caption{
        Reward ablation for Llama-3.
        Bar labels: point-wise changes from Vanilla.
        In the AUROC panels, dotted lines mark the $S^{\text{Det}}$-only ceiling;
        in the task panels, dash-dotted lines mark the $S^{\text{Eval}}$-only ceiling;
        dashed lines mark the Vanilla baseline in all panels.
        ROUGE-L (left) and Judge (right) task axes are proportionally scaled.
    }
    \label{fig:reward_ablation}
\end{figure}

\begin{figure*}[t]
    \centering
    \includegraphics[width=0.9\linewidth]{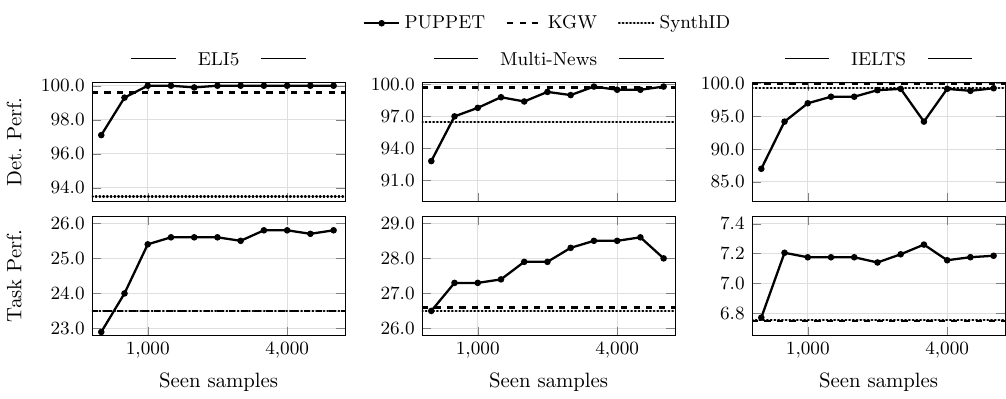}
    \caption{
        Detection performance (Det.\ Perf., AUROC; top row) and task performance
        (Task Perf., ROUGE-L for ELI5 and Multi-News, judge score for IELTS; bottom row)
        as a function of training samples seen for Llama-3 across all three benchmarks
        (500 samples per dot).
        Dashed and dotted lines indicate KGW and SynthID baselines, respectively.
    }
    \label{fig:seen_samples_curve_llama3}
\end{figure*}

PUPPET uses a composite reward comprising a detector score ($S^{\text{Det}}$) and an evaluator score ($S^{\text{Eval}}$).
To assess the necessity of this combined design, we compare PUPPET ($\alpha=0.5$) against single-reward variants: $S^{\text{Det}}$ only ($\alpha\!=\!1$) and $S^{\text{Eval}}$ only ($\alpha\!=\!0$),
which serve as single-objective upper bounds for detectability and task performance, respectively.

As shown in Figure~\ref{fig:reward_ablation}, optimizing detectability alone ($S^{\text{Det}}$ only) achieves large AUROC gains over Vanilla (up to ${+}12.9$ points) but degrades task performance (up to $-1.5$ ROUGE-L and $-0.3$ Judge).
Conversely, optimizing task performance alone ($S^{\text{Eval}}$ only) yields meaningful improvements in it (up to ${+}3.1$ ROUGE-L and ${+}0.8$ Judge) but provides no reliable detection gain.
PUPPET ($\alpha=0.5$) jointly optimizes both objectives: it recovers $95.3$--$100\%$ of the $S^{\text{Det}}$-only detection gain while simultaneously improving task performance over Vanilla.
We additionally evaluate ``$S^{\text{Eval}}$ only ${+}$ KGW''.
Detection rises to the level of KGW ($99.6$, $99.3$, and $100.0$ on ELI5, Multi-News, and IELTS), but task performance falls to that level as well ($23.5$, $26.5$, and $6.76$).
This shows that watermarking largely cancels the task gains of $S^{\text{Eval}}$ only, whereas joint optimization retains them.
These results confirm that both reward components are necessary: neither alone achieves strong performance on both objectives.

\subsection{Training Efficiency} \label{sec:sample_efficiency}

Even a method that achieves strong performance may have limited practical value if it demands prohibitive data or computational resources.
We therefore examine PUPPET's practicality along both dimensions.

\paragraph{Data}
Figure~\ref{fig:seen_samples_curve_llama3} plots detection and task performance as a function of the number of training samples seen for Llama-3.
Across all benchmarks, both detectability and task performance rise sharply within the first few thousand samples and plateau near the level achieved with the full training set.
For example, on ELI5, both performances converge within $1$k--$2$k samples, with AUROC already reaching $100.0$ at $1$k samples.
This handful of DPO pairs represents a remarkably small budget compared to large-scale DPO-based alignment pipelines such as OLMo3~\citep{olmo3}, which train on 150K--200K pairs.
The same trend holds for Qwen3 (\S~\ref{app:practicality_of_training}).
Another data-side cost is the number of candidate responses $k$ sampled per query.
We therefore vary $k \in \{2, 8, 11\}$ in addition to the $k = 5$ used throughout the paper, using IELTS, where both metrics still have headroom (Table~\ref{tab:result_main}).
Task performance is $7.02$, $7.20$, and $7.14$, and detection performance is $94.0$, $99.2$, and $99.3$, respectively, compared with $7.19$ and $99.3$ at $k=5$.
Performance thus improves substantially from $k=2$ to $5$ but saturates thereafter, with $k=8, 11$ offering essentially no further gain.
PUPPET therefore operates effectively with a small candidate pool.

\paragraph{Computational resources}
Building the training data consists of two steps: (1) sampling candidate responses and (2) scoring them,
which take roughly $20$ min and $20$ min on ELI5, $3.5$ h and $20$ min on Multi-News, and $40$ min and $9$ h on IELTS, respectively.
Sampling dominates on Multi-News due to long summarization inputs, and scoring on IELTS due to the LLM-as-a-Judge evaluator.
Each training run then completes in approximately 1--2 hours on a single NVIDIA A6000 (48GB) GPU, so the entire pipeline fits within $12$ hours in every setting.
Moreover, because PUPPET is based on offline DPO, these costs are paid only once: sampling once per dataset--model pair, and scoring once each for the detector and the evaluator.
Reconfiguring PUPPET then reduces to retraining; changing $\alpha$, for instance, only recombines the cached scores into new preference pairs.

These results confirm that PUPPET is both data-efficient and computationally lightweight.

\begin{table}[t]
    \setlength{\tabcolsep}{1.5mm}
    \centering
    \small
    \begin{tabular}{llccc}
        \toprule
        \multicolumn{2}{l}{\texttt{Cross-Detector}} & \textbf{ELI5} & \textbf{MN} & \textbf{IELTS} \\
        \cmidrule(lr){3-3} \cmidrule(lr){4-4} \cmidrule(lr){5-5}
        \textbf{Detector} & \textbf{Method} & \textbf{AUC} & \textbf{AUC} & \textbf{AUC} \\
        \midrule

        \multicolumn{5}{c}{\textsc{RoBERTa-based classifier}} \\
        \midrule
        
        \multirow{2}{*}{\makecell[l]{FakeSpotAI}}
         & \graycell{Vanilla} & \graycell{100.0} & \graycell{100.0} & \graycell{99.7} \\
         & PUPPET \puppeticon & \best{100.0} & \best{100.0} & \best{99.8} \\
        \graymidrule
        
        \multirow{2}{*}{MAGE}
         & \graycell{Vanilla} & \graycell{100.0} & \graycell{\best{97.2}} & \graycell{\best{89.4}} \\
         & PUPPET \puppeticon & \best{100.0} & 96.9 & 87.7 \\
        \graymidrule
        
        \multirow{2}{*}{RADAR}
         & \graycell{Vanilla} & \graycell{97.7} & \graycell{99.3} & \graycell{94.3} \\
         & PUPPET \puppeticon & \best{99.5} & \best{99.5} & \best{97.2} \\

        \midrule
        \multicolumn{5}{c}{\textsc{Zero-shot detector}} \\
        \midrule
        
        \multirow{2}{*}{\makecell[l]{Fast-DetectGPT}}
         & \graycell{Vanilla} & \graycell{100.0} & \graycell{99.0} & \graycell{100.0} \\
         & PUPPET \puppeticon & \best{100.0} & \best{99.8} & \best{100.0} \\
        \graymidrule

        \multirow{2}{*}{Binoculars}
         & \graycell{Vanilla} & \graycell{100.0} & \graycell{99.9} & \graycell{100.0} \\
         & PUPPET \puppeticon & \best{100.0} & \best{100.0} & \best{100.0} \\
        
        \bottomrule
    \end{tabular}
    \caption{
        Cross-detector evaluation.
        All PUPPET models are fine-tuned with the OpenAI detector, and evaluated with the other detectors.
        Bold: best per detector for each task.
    }
    \label{tab:cross_detector}
\end{table}

\subsection{Generalizability to Unseen Detectors} \label{sec:cross_detectors}
As a supplementary analysis, we evaluate PUPPET in the \textbf{cross-detector} setting.
Specifically, for each task, we take the PUPPET model fine-tuned with the OpenAI detector and evaluate it with the five other detectors used in \S~\ref{sec:robustness_detector}.
Note that, as stated in \S~\ref{sec:intro}, PUPPET is designed to improve detectability with respect to a target detector.
This setting thus lies outside its intended use. Table~\ref{tab:cross_detector} reports the results.

Compared with the multi-detector setting (\S~\ref{sec:robustness_detector}), where each detector is also used for fine-tuning,
the average change per detector is smaller or equal for all five detectors:
for instance (\textit{multi}$\to$\textit{cross}), ${+}1.0\to{-}0.7$ for MAGE and ${+}2.2\to{+}1.6$ for RADAR.
Overall, the gains from PUPPET are largely specific to the detector used during fine-tuning.

\subsection{PUPPET Potentially Enables Model Attribution} \label{sec:model_attribution}

As an exploratory analysis, we investigate the use of PUPPET for model attribution.
Since PUPPET raises the machine-generated class likelihood of the target model, a sufficiently high threshold on this likelihood should separate that model from models without PUPPET applied.
We evaluate this by treating \textit{Target} model outputs as the positive (machine) class and \textit{Contrast} model outputs as the negative (human) class, computing AUROC from the OpenAI Detector's machine-generated class likelihood.Table~\ref{tab:distinguish_models} shows Llama-3 results.

Attribution accuracy (AUROC) is governed by the gap in machine-generated class likelihood between the two models; we quantify this with Cohen's $d$, reported alongside AUROC in Table~\ref{tab:distinguish_models}.
At the Vanilla baseline, the gap is small on ELI5 ($d = 0.23$), yielding near-chance attribution (AUROC $58.99$), and grows on Multi-News and IELTS, supporting only moderate performance ($72.26$--$77.71$).
After PUPPET, the gap widens substantially (up to $d = 4.62$ on IELTS) and attribution rises markedly to $94.47$--$99.14$, confirming that PUPPET's controlled upward shift creates a reliable threshold.
This suggests that a publisher of Llama-3 who applies PUPPET before release could, after release, tell its own model's outputs apart from those of other models.

We treat these findings as preliminary evidence of PUPPET's potential for model attribution.
The similar trend holds for Qwen3 (see \S~\ref{app:model_attribution}).

\begin{table}[t]
    \centering
    \small
    \setlength{\tabcolsep}{1mm}
    \begin{tabular}{llccc}
    \toprule
    \textbf{Target} & \textbf{Contrast}
      & \textbf{ELI5} & \textbf{Multi-News} & \textbf{IELTS} \\
    \midrule
    \rowcolor{lightgrayrow}
    Llama-3 & Qwen3   & 58.99 {\scriptsize(0.23)} & 72.26 {\scriptsize(0.82)} & 77.71 {\scriptsize(0.96)} \\
    Llama-3\puppeticon  & Qwen3   & \best{94.47} {\scriptsize(1.77)} & \best{96.94} {\scriptsize(3.18)} & \best{99.14} {\scriptsize(4.62)} \\
    \bottomrule
    \end{tabular}
    \caption{
      Model-attribution AUROC (OpenAI Detector), treating \textit{Target} as the positive class and \textit{Contrast} as the negative.
      Parentheses: Cohen's $d$; models are trained per task.
    }
    \label{tab:distinguish_models}
\end{table}

\section{Related Work} \label{sec:related_works}
Research on machine text detection can be broadly categorized into two directions:
(1) detectors: using discriminative features of machine-generated text, and
(2) watermarking: embedding detectable features into text during generation.

\paragraph{Detectors}
A standard approach involves training supervised classifiers on labeled datasets of human-written and machine-generated text.
RoBERTa~\citep{roberta} is the most widely used base model~\citep{roberta_yuxia, roberta-tulchinskii}:
the OpenAI Detector~\citep{openai_detector} serves as a common benchmark in prior work~\citep{easily_attacked, roft}, MAGE~\citep{roberta-yafu} is trained on diverse in-the-wild domains and generators,
RADAR~\citep{radar} is trained adversarially against a paraphraser, and FakeSpotAI is one of ApolloDFT models designed for detecting deepfakes.

A complementary, training-free paradigm exploits statistical properties of the generation process.
DetectGPT~\citep{detectgpt} exploits the observation that machine-generated text lies near local maxima of a model’s log-probability surface, so perturbations lower its log-probability, whereas human-written text does not exhibit this behavior.
Fast-DetectGPT~\citep{fastdetectgpt} achieves comparable accuracy at substantially lower cost by replacing perturbation-based estimation with conditional probability curvature sampling.
Binoculars~\citep{binoculars} achieves strong zero-shot detection by computing the ratio of cross-perplexity and perplexity between two related models.

\paragraph{Watermarking}
Watermarking methods embed detectable signals into LLM-generated text and can be grouped into logit-based and sampling-based approaches.

\textit{Logit-based methods} bias the token output distribution at generation time.
KGW~\citep{kgw} partitions the vocabulary into green and red lists conditioned on the preceding token, and increases the logits of green-list tokens by a fixed $\delta$.
Unigram~\citep{unigram} simplifies this by using a fixed, context-independent green list, thereby making it more robust to token-substitution attacks.
Both methods introduce systematic distortions into the output distribution, which has been shown to degrade task performance~\citep{tradeoff_ajith, tradeoff_yu, tradeoff_wang, tradeoff_fernandez}.

\textit{Sampling-based methods}, in contrast, intervene at the token selection stage without directly modifying logits.
EXPGumbel~\citep{expgumbel} resamples tokens via a seeded Gumbel distribution, and SynthID~\citep{synthid} selects tokens through a tournament procedure over multiple random scores.
K-SemStamp~\citep{ksemstamp} is a clustering-based extension of SemStamp~\citep{semstamp},
which steers sentence-level generation toward watermarked regions in semantic embedding space.
These methods are generally less disruptive to the output distribution than logit-based approaches, but task performance degradation can still be observed~\citep{tradeoff_wang}.

\section{Conclusion} \label{sec:conclusion}
We presented PUPPET, the DPO-based framework that fine-tunes an LLM to produce text that is both highly detectable and performant on downstream tasks.
By such joint optimization, PUPPET matches watermarking methods in detection while surpassing them on downstream tasks.
These gains generalize to out-of-domain tasks and prove robust to the choice of base LLM, detector, and paraphrasing attacks, all within moderate data and computational resources.
Our exploratory analysis further suggests that PUPPET opens the door to model attribution.
We hope this work encourages research into jointly optimizing LLMs across multiple accountability objectives, thereby contributing to more transparent deployment and fair use of language models.

\section*{Ethical Statement} 
Detection approaches, including PUPPET and watermarking, are still in the research stage.
Therefore, it is essential to use these techniques with utmost caution under high-stakes settings where incorrectly identifying human-written text as machine-generated text may cause significant harm.

\bibliography{aaai2027}

\appendix
\section{Implementation Details} \label{app:details}

\subsection{Generation Settings} \label{app:generation_setup}

All text generation---both for constructing DPO preference data and for evaluation---uses vLLM~\citep{vllm} (v0.11.0).
To evaluate post-trained models correctly, we apply chat templates.
All parameters related to generation use defaults, except for \texttt{n} (number of responses per prompt; see \S~\ref{sec:setup}) and \texttt{enable\_thinking} (\texttt{False} for all Qwen3 experiments, to exclude any effects of thinking).

\subsection{Training Setup} \label{app:training_setup}

We fine-tune with the \texttt{DPOTrainer} in TRL\footnote{\url{https://github.com/huggingface/trl}}, using LoRA for memory efficiency.
Table~\ref{tab:hparams} lists the key hyperparameters; values not shown use defaults.
We adopt a single fixed configuration without systematic search except for the ablation studies of $\alpha$ and $k$ in \S~\ref{sec:analysis}.

\subsection{Evaluation Setup} \label{app:evaluation_setup}

\subsubsection{ROUGE-L}
ROUGE-L for ELI5 and Multi-News follows the WaterBench~\citep{waterbench} implementation, where we compute the ROUGE-L score for each response and adopt the best score per sample.
Since ROUGE-L captures only surface-level lexical overlap, we additionally re-evaluate Llama-3 with BARTScore~\citep{bartscore}, a model-based semantic metric, and confirm that PUPPET remains the best on both tasks (Table~\ref{tab:bartscore}).

\subsubsection{LLM-as-a-Judge Evaluation} \label{app:judge}
For IELTS, we use an LLM-as-a-Judge score on a $[0, 9]$ IELTS band scale as a task performance metric.
We use GPT-OSS-20B as the judge with temperature $0.0$ and \texttt{max\_tokens}~$= 4096$.
The scoring prompt is adapted from \citet{essay_prompt}.

\textit{Probability-weighted scoring.} ---
Rather than parsing the generated integer directly, we compute the final score as a probability-weighted expected value over the ten score labels $\{0, 1, \ldots, 9\}$, following \citet{geval}:
\begin{equation*}
    \hat{s} \;=\; \sum_{v=0}^{9}\; v \cdot P(v),
    \qquad
    P(v) \;=\; \frac{\exp(\ell_v)}{\displaystyle\sum_{v'=0}^{9} \exp(\ell_{v'})},
\end{equation*}
where $\ell_v$ is the log-probability of score token $v$ as assigned by the judge model.
This yields a continuous score in $[0, 9]$ and, crucially,
produces sufficient variance across the $k = 5$ preference candidates---a necessary condition for effective DPO training (detailed at the ablation study of $k$ in \S~\ref{sec:sample_efficiency}).

\begin{table}[t]
    \centering
    \small
    \begin{tabular}{lcc}
        \toprule
        \textbf{Method} & \textbf{ELI5} & \textbf{Multi-News} \\
        \midrule
        KGW & $-3.66$ & $-2.96$ \\
        \rowcolor{lightgrayrow}
        \midrule
        Vanilla & $-3.72$ & $-2.98$ \\
        \textbf{PUPPET\puppeticon} & $\mathbf{-3.54}$ & $\mathbf{-2.86}$ \\
        \bottomrule
    \end{tabular}
    \caption{
        BARTScore (higher is better) on ELI5 and Multi-News for Llama-3.
        Bold: best per column.
        Vanilla denotes the base model without PUPPET fine-tuning.
    }
    \label{tab:bartscore}
\end{table}

\begin{table}[t]
    \centering
    \small
    \begin{tabular}{ll}
        \toprule
        \textbf{Hyperparameter} & \textbf{Value} \\
        \midrule
        Learning rate         & $1 \times 10^{-4}$ \\
        Epochs                & 1 \\
        Batch size (per device) & ELI5: 20; MN: 5; IELTS: 10 \\
        Data seed             & 42 \\
        Max sequence length  & ELI5: 800; MN / IELTS: 8{,}192 \\
        Max completion length & ELI5: 300; MN / IELTS: 512 \\
        \bottomrule
    \end{tabular}
    \caption{
    Training parameters we modified from the defaults.
    }
    \label{tab:hparams}
\end{table}

\begin{table}[!t]
    \centering
    \small
    \setlength{\tabcolsep}{1mm}
    \begin{tabular}{ll>{\raggedright\arraybackslash}p{3cm}r}
        \toprule
        \textbf{Task} & \textbf{Split} & \textbf{Source} & \textbf{Samples} \\
        \midrule
        \multirow{2}{*}{ELI5}
         & Train & Hello-SimpleAI/HC3, reddit\_eli5 & 5{,}000 \\
         & Eval  & THU-KEG/WaterBench (2-1\_longform\_qa) & 200 \\
        \midrule
        \multirow{2}{*}{Multi-News}
         & Train & alexfabbri/multi\_news & 5{,}000 \\
         & Eval  & THU-KEG/WaterBench (4-1\_multi\_news)& 200 \\
        \midrule
        \multirow{2}{*}{IELTS}
         & Train & chillies/IELTS-writing-task-2-evaluation & 5{,}000 \\
         & Eval  & chillies/IELTS-writing-task-2-evaluation & 200 \\
        \bottomrule
    \end{tabular}
    \caption{
    Data sources for training and evaluation splits.
    }
    \label{tab:datasets}
\end{table}

\paragraph{Handling the reasoning model.}
Because GPT-OSS-20B is a reasoning model, it generates a thinking trajectory before producing the final score token.
To compute token-level log-probabilities conditioned on that reasoning, we proceed in two steps:
(1) we first generate the full thinking trajectory by sampling with a stop sequence that marks the end of the thinking phase;
(2) we then prepend the thinking trajectory and the prefix
``\texttt{Score:~}'' to each candidate score token and extract the corresponding prompt log-probability via vLLM.
This ensures that $\ell_v$ reflects the model's judgment after reasoning rather than being estimated unconditionally.
Step (1) may fail to complete the thinking trajectory within the token limit (\texttt{max\_new\_tokens}).
We set the maximum number of retries to 20 and exclude samples whose thinking trajectory fails to terminate.\footnote{
LLM-as-a-Judge failures were negligible: 3 out of 25{,}000 judgments (5{,}000 samples $\times$ 5 responses) for Llama-3 and
1 out of 25{,}000 for Qwen3. The corresponding samples were excluded from preference pair construction.
}

\subsection{Dataset} \label{app:dataset}
The training and evaluation datasets are summarized in Table~\ref{tab:datasets}.
We use $42$ as the random seed for all data sampling.

\subsection{Watermark Baseline Setup} \label{app:watermark_setup}
We use MarkLLM~\citep{markllm} to generate all watermarked texts (KGW, Unigram, EXPGumbel, SynthID, K-SemStamp),
with the only modification being the application of chat templates for better evaluation of post-trained models.
We adopt the default configurations bundled in MarkLLM.

\subsection{Span Visualization} \label{app:shap}

\begin{table}[t]
    \centering
    \setlength{\tabcolsep}{1mm}
    \small
    \begin{tabular}{lcccccc}
        \toprule
        \texttt{Qwen3-8B}
        & \multicolumn{2}{c}{\textbf{ELI5}}
        & \multicolumn{2}{c}{\textbf{Multi-News}}
        & \multicolumn{2}{c}{\textbf{IELTS}} \\
        \cmidrule(lr){2-3}\cmidrule(lr){4-5}\cmidrule(lr){6-7}
        \textbf{Method} & \textbf{AUC} & \textbf{R-L} & \textbf{AUC} & \textbf{R-L} & \textbf{AUC} & \textbf{Jdg.} \\
        \midrule
        \multicolumn{7}{c}{\textsc{Watermarking Baselines}} \\
        \midrule
        KGW        & 99.4 & 24.5 & \best{99.3} & 24.1 & \best{99.9} & \best{7.78}\sigup \\
        SynthID    & 94.5 & 24.6 & 98.0 & 24.2 & 98.9 & 7.67 \\
        Unigram    & 91.1 & 24.1\sigdown & 97.0 & 24.3 & 99.1 & 7.68 \\
        EXPGumbel  & 98.8 & 23.7\sigdown & 96.3 & 23.7 & 97.8 & 4.89\sigdown\\
        K-SemStamp & 98.6 & 24.7 & 98.5 & 23.4\sigdown & 98.5 & 7.67 \\
        \midrule
    
        \multicolumn{7}{c}{\textsc{Ours}} \\
        \midrule
        \rowcolor{lightgrayrow}
        Vanilla & 93.7 & 24.5 & 83.3 & 24.0 & 66.6 & 7.63 \\
        \textbf{PUPPET\puppeticon} & \best{99.9} & \best{26.6}\sigup & 96.7 & \best{26.4}\sigup & 85.6 & 7.65 \\
        \bottomrule
    \end{tabular}
    \caption{
        In-domain detection and task performance for Qwen3-8B.
        Bold: best per column.
        \sigupcap{} and \sigdowncap{} denote significant improvement and degradation
        on task performance (paired $t$-test, $\alpha=0.05$), respectively.
    }
    \label{tab:result_main_qwen_indomain}
\end{table}

\begin{table}
    \setlength{\tabcolsep}{1mm}
    \centering
    \small
    \begin{NiceTabular}{lcccccc}
        \toprule
        \texttt{Qwen3-8B}
        & \multicolumn{2}{c}{\textbf{ELI5}}
        & \multicolumn{2}{c}{\textbf{Multi-News}}
        & \multicolumn{2}{c}{\textbf{IELTS}} \\
        \cmidrule(lr){2-3}\cmidrule(lr){4-5}\cmidrule(lr){6-7}
        \textbf{Train}
        & \textbf{AUC} & \textbf{R-L}
        & \textbf{AUC} & \textbf{R-L}
        & \textbf{AUC} & \textbf{Jdg.} \\
        \midrule

        \rowcolor{lightgrayrow}
        Vanilla
        & 93.7 & 24.5
        & 83.3 & 24.0
        & 66.6 & 7.62 \\

        ELI5 \puppeticon
        & \best{99.9} & \best{26.6}\sigup
        & 90.1 & 24.4
        & 82.1 & \best{7.65} \\

        Multi-News \puppeticon
        & 98.6 & 25.4\sigup
        & \best{96.7} & \best{26.4}\sigup
        & \best{87.4} & 7.52 \\
        
        IELTS \puppeticon
        & 98.5 & 25.3\sigup
        & 89.0 & 24.6\sigup
        & 85.6 & \best{7.65} \\
        
        \bottomrule
        
        \CodeAfter
        \begin{tikzpicture}[overlay]
        \def\xpad{2.0pt}
        \def\ypadtop{1.6pt}
        \def\ypadbot{0.8pt}
        \def\ynotch{1.2pt}
        
        \draw[line width=0.4pt, rounded corners=2pt]
            ([xshift=-\xpad,yshift=\ypadtop]4-4.north west) --
            ([xshift=\xpad,yshift=\ypadtop]4-7.north east) --
            ([xshift=\xpad,yshift=-\ypadbot]5-7.south east) --
            ([xshift=-\xpad,yshift=-\ypadbot]5-6.south west) --
            ([xshift=-\xpad,yshift=\ypadtop]5-6.north west) --
            ([xshift=-\xpad,yshift=-\ypadbot]4-4.south west) -- cycle;
        
        \draw[line width=0.4pt, rounded corners=2pt]
            ([xshift=-\xpad,yshift=\ypadtop]5-2.north west) --
            ([xshift=\xpad,yshift=\ypadtop]5-3.north east) --
            ([xshift=\xpad,yshift=\ynotch]6-3.north east) --
            ([xshift=\xpad,yshift=\ynotch]6-5.north east) --
            ([xshift=\xpad,yshift=-\ypadbot]6-5.south east) --
            ([xshift=-\xpad,yshift=-\ypadbot]6-2.south west) --
            ([xshift=-\xpad,yshift=-\ypadbot]5-2.south west) -- cycle;
        \end{tikzpicture}
    \end{NiceTabular}
    \caption{
        Detection and task performance for Qwen3-8B.
        Rows: task used for fine-tuning; columns: evaluation benchmarks.
        Boxes highlight OOD evaluations (off-diagonal).
        \sigupcap{} and \sigdowncap{} denote significant improvement and degeneration on task performance (paired $t$-test, $\alpha=0.05$), respectively.
    }
    \label{tab:robustness_ood_qwen}
\end{table}

The span visualization with Llama-3 on ELI5 (Figure~\ref{fig:puppet_shap}) is produced using the Python \texttt{shap} library\footnote{\url{https://github.com/maciejskorski/shap}}, operating at the granularity of the model's tokenizer.

\paragraph{Detection-relevant spans (red).}
We compute SHAP values for the OpenAI Detector with respect to class 0 (Machine).
Only the positive component of each token's SHAP value is retained (\texttt{mode = "pos"}), capturing tokens that actively push the prediction toward the Machine class.
To normalize scores comparably across conditions, we compute a shared upper bound $c_{\max}$ as the 99th percentile of the pooled positive SHAP scores from both the Vanilla and PUPPET outputs, ensuring that the color scale is directly comparable between the two conditions.
Tokens whose normalized score meets or exceeds the threshold $\tau = 0.4$ are grouped into contiguous spans and highlighted in red.

\paragraph{Task-relevant spans (blue).}
For ROUGE-L, task-relevant spans are identified via the longest common subsequence (LCS) between the generated text and the reference answer, computed at the word level (whitespace tokenization).
Matched word spans are mapped back to character offsets and highlighted in blue.
Extending this visualization to LLM-as-a-Judge spans proved computationally prohibitive ($>$1 week), and we therefore report only ROUGE-L-based spans.

\subsection{Paraphrasing Attack Setup} \label{app:paraphrase_setup}
To evaluate robustness to adversarial paraphrasing, we apply Dipper~\citep{dipper} to all model outputs before running detection.
All experiments use identical Dipper settings across benchmarks and base models. 
We set \texttt{lex\_diversity} and \texttt{order\_diversity} to 60, following the strongest parameter values reported by \citet{dipper}.
Other hyperparameters use defaults.

\begin{table}[t]
    \centering
    \setlength{\tabcolsep}{1mm}
    \small
    \begin{tabular}{lcccccc}
        \toprule
        \texttt{Qwen3-4B}
        & \multicolumn{2}{c}{\textbf{ELI5}}
        & \multicolumn{2}{c}{\textbf{Multi-News}}
        & \multicolumn{2}{c}{\textbf{IELTS}} \\
        \cmidrule(lr){2-3}\cmidrule(lr){4-5}\cmidrule(lr){6-7}
        \textbf{Method} & \textbf{AUC} & \textbf{R-L} & \textbf{AUC} & \textbf{R-L} & \textbf{AUC} & \textbf{Jdg.} \\
        \midrule
        \multicolumn{7}{c}{\textsc{Watermarking Baselines}} \\
        \midrule
        KGW        & 96.2 & 24.3 & \best{96.4} & 25.2 & \best{99.5} & 7.57 \\
        SynthID    & 86.8 & 24.5 & 93.5 & 25.1 & 97.3 & 7.60 \\
        Unigram    & 73.5 & 24.3 & 85.3 & 25.4\sigup & 97.0 & \best{7.63}\sigup \\
        EXPGumbel  & 88.4 & 23.8\sigdown & 87.9 & 25.4 & 95.9 & 4.86\sigdown \\
        K-SemStamp & 96.1 & 24.3 & 95.4 & 24.2\sigdown & 98.2 & 7.57 \\
    
        \midrule
        \multicolumn{7}{c}{\textsc{Ours}} \\
        \midrule
        \rowcolor{lightgrayrow}
        Vanilla & 98.2 & 24.5 & 88.0 & 24.9 & 79.9 & 7.50 \\
        \textbf{PUPPET\puppeticon} & \best{100.0} & \best{25.6}\sigup & 95.4 & \best{26.7}\sigup & 96.0 & 7.53 \\
        \bottomrule
    \end{tabular}
    \caption{
        In-domain detection and task performance for Qwen3-4B.
        Bold: best per column.
        \sigupcap{} and \sigdowncap{} denote significant improvement and degradation
        on task performance (paired $t$-test, $\alpha=0.05$), respectively.
    }
    \label{tab:result_main_qwen4b_indomain}
\end{table}

\begin{table}[t]
    \setlength{\tabcolsep}{1mm}
    \centering
    \small
    \begin{NiceTabular}{lcccccc}
        \toprule
        \texttt{Qwen3-4B}
         & \multicolumn{2}{c}{\textbf{ELI5}}
         & \multicolumn{2}{c}{\textbf{Multi-News}}
         & \multicolumn{2}{c}{\textbf{IELTS}} \\
        \cmidrule(lr){2-3}\cmidrule(lr){4-5}\cmidrule(lr){6-7}
        \textbf{Train}
         & \textbf{AUC} & \textbf{R-L}
         & \textbf{AUC} & \textbf{R-L}
         & \textbf{AUC} & \textbf{Jdg.} \\
        \midrule

        \rowcolor{lightgrayrow}
        Vanilla
         & 98.2 & 24.5
         & 88.0 & 24.9
         & 79.9 & 7.50 \\

        ELI5 \puppeticon
         & \best{100.0} & \best{25.6}\sigup
         & 94.4 & 25.6\sigup
         & 89.7 & \best{7.39} \\

        Multi-News \puppeticon
         & 99.0 & 25.2\sigup
         & \best{95.4} & \best{26.7}\sigup
         & 89.2 & 7.45 \\
        
        IELTS \puppeticon
         & 99.8 & 24.8\sigup
         & 95.6 & 25.6\sigup
         & \best{96.0} & \best{7.53} \\
        
        \bottomrule
        
        \CodeAfter
        \begin{tikzpicture}[overlay]
          \def\xpad{2.0pt}
          \def\ypadtop{1.6pt}
          \def\ypadbot{0.8pt}
          \def\ynotch{1.2pt}
        
          \draw[line width=0.4pt, rounded corners=2pt]
            ([xshift=-\xpad,yshift=\ypadtop]4-4.north west) --
            ([xshift=\xpad,yshift=\ypadtop]4-7.north east) --
            ([xshift=\xpad,yshift=-\ypadbot]5-7.south east) --
            ([xshift=-\xpad,yshift=-\ypadbot]5-6.south west) --
            ([xshift=-\xpad,yshift=\ypadtop]5-6.north west) --
            ([xshift=-\xpad,yshift=-\ypadbot]4-4.south west) -- cycle;
        
          \draw[line width=0.4pt, rounded corners=2pt]
            ([xshift=-\xpad,yshift=\ypadtop]5-2.north west) --
            ([xshift=\xpad,yshift=\ypadtop]5-3.north east) --
            ([xshift=\xpad,yshift=\ynotch]6-3.north east) --
            ([xshift=\xpad,yshift=\ynotch]6-5.north east) --
            ([xshift=\xpad,yshift=-\ypadbot]6-5.south east) --
            ([xshift=-\xpad,yshift=-\ypadbot]6-2.south west) --
            ([xshift=-\xpad,yshift=-\ypadbot]5-2.south west) -- cycle;
        \end{tikzpicture}
    \end{NiceTabular}
    \caption{
    Detection and task performance for Qwen3-4B.
    Rows: task used for fine-tuning; columns: evaluation benchmarks.
    Boxes highlight OOD evaluations (off-diagonal).
    \sigupcap{} and \sigdowncap{} denote significant improvement and degeneration on task performance (paired $t$-test, $\alpha=0.05$), respectively.
    }
    \label{tab:robustness_ood_qwen4b}
\end{table}

\section{Full Results for Other Base LLMs} \label{app:qwen}
This appendix provides full per-task results for all Qwen3 models evaluated in \S~\ref{sec:robustness}.
organized by the robustness axis under investigation:
\S~\ref{app:qwen_family} addresses robustness to model family (Qwen3-8B vs.\ Llama-3-8B-Instruct),
and \S~\ref{app:qwen_size} addresses robustness to model size (Qwen3-4B, 8B, and 14B).

\subsection{Model Family: Qwen3-8B} \label{app:qwen_family}

Table~\ref{tab:result_main_qwen_indomain} shows ID results for Qwen3-8B.
On ELI5 and Multi-News, PUPPET markedly improves both detection and task performance compared to Vanilla, consistent with the Llama-3 results in Table~\ref{tab:result_main}.
On IELTS, however, the AUROC gain is more modest ($66.6 \to 85.6$) compared to Llama-3 ($87.0 \to 99.3$).
The lower Vanilla AUROC on IELTS for Qwen3-8B suggests that its essay-style outputs are further from the OpenAI Detector's training distribution, making the detection signal harder to amplify via DPO.
Despite this, PUPPET still achieves a meaningful absolute gain, confirming that PUPPET is not inherently limited to a Llama3 family.

Table~\ref{tab:robustness_ood_qwen} shows OOD results.
As with Llama-3 (Table~\ref{tab:robustness_ood}), PUPPET consistently improves detection performance over Vanilla even on tasks not seen during training, confirming that the detection-relevant features learned by PUPPET transfer across tasks regardless of the base model family.

\subsection{Model Size: Qwen3-4B and 14B} \label{app:qwen_size}

\begin{table}[t]
    \centering
    \setlength{\tabcolsep}{1mm}
    \small
    \begin{tabular}{lcccccc}
        \toprule
        \texttt{Qwen3-14B}
        & \multicolumn{2}{c}{\textbf{ELI5}}
        & \multicolumn{2}{c}{\textbf{Multi-News}}
        & \multicolumn{2}{c}{\textbf{IELTS}} \\
        \cmidrule(lr){2-3}\cmidrule(lr){4-5}\cmidrule(lr){6-7}
        \textbf{Method} & \textbf{AUC} & \textbf{R-L} & \textbf{AUC} & \textbf{R-L} & \textbf{AUC} & \textbf{Jdg.} \\
        \midrule
        \multicolumn{7}{c}{\textsc{Watermarking Baselines}} \\
        \midrule
        KGW        & 99.2 & 24.4 & \best{99.7} & 24.3 & \best{99.8} & \best{7.88} \\
        SynthID    & 94.5 & 24.2 & 98.1 & 24.4 & 99.3 & 7.80 \\
        Unigram    & 90.7 & 24.0\sigdown & 96.8 & 24.3 & 98.6 & 7.78 \\
        EXPGumbel  & 98.5 & 23.3\sigdown & 96.9 & 24.3 & 98.1 & 4.98\sigdown \\
        K-SemStamp & 98.0 & 24.2 & 99.0 & 23.2\sigdown & 99.0 & 7.84 \\
        
        \midrule
        \multicolumn{7}{c}{\textsc{Ours}} \\
        \midrule
        \rowcolor{lightgrayrow}
        Vanilla & 86.4 & 24.4 & 76.4 & 24.0 & 65.1 & 7.86 \\
        \textbf{PUPPET\puppeticon} & \best{99.6} & \best{26.0}\sigup & 99.1 & \best{28.3}\sigup & 93.6 & 7.81 \\
        \bottomrule
    \end{tabular}
    \caption{
        In-domain detection and task performance for Qwen3-14B.
        Bold: best per column.
        \sigupcap{} and \sigdowncap{} denote significant improvement and degradation
        on task performance (paired $t$-test, $\alpha=0.05$), respectively.
    }
    \label{tab:result_main_qwen14b_indomain}
\end{table}

\begin{table}[t]
    \setlength{\tabcolsep}{1mm}
    \centering
    \small
    \begin{NiceTabular}{lcccccc}
        \toprule
        \texttt{Qwen3-14B}
         & \multicolumn{2}{c}{\textbf{ELI5}}
         & \multicolumn{2}{c}{\textbf{Multi-News}}
         & \multicolumn{2}{c}{\textbf{IELTS}} \\
        \cmidrule(lr){2-3}\cmidrule(lr){4-5}\cmidrule(lr){6-7}
        \textbf{Train}
         & \textbf{AUC} & \textbf{R-L}
         & \textbf{AUC} & \textbf{R-L}
         & \textbf{AUC} & \textbf{Jdg.} \\
        \midrule

        \rowcolor{lightgrayrow}
        Vanilla
         & 86.4 & 24.4
         & 76.4 & 24.0
         & 65.1 & \best{7.86} \\

        ELI5 \puppeticon
         & \best{99.6} & \best{26.0}\sigup
         & 82.3 & 24.7\sigup
         & 80.6 & 7.73\sigdown \\

        Multi-News \puppeticon
         & 98.5 & 25.4\sigup
         & \best{99.1} & \best{28.3}\sigup
         & 93.4 & 7.44\sigdown \\
        
        IELTS \puppeticon
         & 97.2 & 25.0\sigup
         & 86.6 & 24.8\sigup
         & \best{93.6} & 7.81 \\
        
        \bottomrule
        
        \CodeAfter
        \begin{tikzpicture}[overlay]
          \def\xpad{2.0pt}
          \def\ypadtop{1.6pt}
          \def\ypadbot{0.8pt}
          \def\ynotch{1.2pt}
        
          \draw[line width=0.4pt, rounded corners=2pt]
            ([xshift=-\xpad,yshift=\ypadtop]4-4.north west) --
            ([xshift=\xpad,yshift=\ypadtop]4-7.north east) --
            ([xshift=\xpad,yshift=-\ypadbot]5-7.south east) --
            ([xshift=-\xpad,yshift=-\ypadbot]5-6.south west) --
            ([xshift=-\xpad,yshift=\ypadtop]5-6.north west) --
            ([xshift=-\xpad,yshift=-\ypadbot]4-4.south west) -- cycle;
        
          \draw[line width=0.4pt, rounded corners=2pt]
            ([xshift=-\xpad,yshift=\ypadtop]5-2.north west) --
            ([xshift=\xpad,yshift=\ypadtop]5-3.north east) --
            ([xshift=\xpad,yshift=\ynotch]6-3.north east) --
            ([xshift=\xpad,yshift=\ynotch]6-5.north east) --
            ([xshift=\xpad,yshift=-\ypadbot]6-5.south east) --
            ([xshift=-\xpad,yshift=-\ypadbot]6-2.south west) --
            ([xshift=-\xpad,yshift=-\ypadbot]5-2.south west) -- cycle;
        \end{tikzpicture}
    \end{NiceTabular}
    \caption{
    Detection and task performance for Qwen3-14B.
    Rows: task used for fine-tuning; columns: evaluation benchmarks.
    Boxes highlight OOD evaluations (off-diagonal).
    \sigupcap{} and \sigdowncap{} denote significant improvement and degeneration on task performance (paired $t$-test, $\alpha=0.05$), respectively.
    }
    \label{tab:robustness_ood_qwen14b}
\end{table}

Tables~\ref{tab:result_main_qwen4b_indomain} and~\ref{tab:robustness_ood_qwen4b} report results for Qwen3-4B,
and Tables~\ref{tab:result_main_qwen14b_indomain} and~\ref{tab:robustness_ood_qwen14b} report results for Qwen3-14B.

\paragraph{Detection performance.}
As observed for Llama-3 in Table~\ref{tab:result_main},
PUPPET's detection AUROC is competitive with watermarking baselines on both Qwen3-4B and Qwen3-14B,
while consistently surpassing them on task performance.

\paragraph{Task performance.}
Task performance improves consistently across both sizes.
For Qwen3-14B, ROUGE-L improve substantially on both ELI5 and Multi-News ($24.4 \to 26.0$ and $24.1 \to 28.3$, respectively), while Judge score on IELTS decreases marginally ($7.86 \to 7.81$).
However, the degradation is not statistically significant and is small relative to the detection gain ($65.1 \to 93.6$).

\paragraph{Out-of-domain generalization.}
The OOD results show that cross-task generalization holds for both Qwen3-4B and 14B.
PUPPET models fine-tuned on any single task consistently improves detection on the other two tasks over Vanilla.

\section{OOD Results: FakeSpotAI and MAGE} \label{app:detectors}

Among the alternative detectors evaluated in \S~\ref{sec:robustness},
FakeSpotAI and MAGE are the two for which PUPPET does not improve detection and task performance simultaneously on Multi-News in domain (Table~\ref{tab:robustness_detector}).
This appendix examines whether that behavior also carries over to the OOD setting.
Table~\ref{tab:multi_detector_ood} reports the results.

Cross-task generalization remains strong for both detectors and both base models, as observed for the OpenAI detector (Table~\ref{tab:robustness_ood}).
For FakeSpotAI, every off-diagonal AUROC is at least $99.4$ and never falls below Vanilla.
For MAGE, off-diagonal AUROC matches or exceeds Vanilla in $8$ of the $12$ cases, with the remaining drops at most $0.5$ points.
The in-domain behavior that motivated this analysis thus does not propagate out of domain:
MAGE's ${-}1.8$ AUROC drop is confined to the diagonal, where it coincides with its largest ROUGE-L gain ($26.5 \to 29.2$),
suggesting a mild trade-off between the two objectives.

\begin{table}[t]
    \setlength{\tabcolsep}{1mm}
    \centering
    \small
    \begin{NiceTabular}{ll cccccc}
        \toprule
        \textbf{Detector} & \textbf{Train}
         & \multicolumn{2}{c}{\textbf{ELI5}}
         & \multicolumn{2}{c}{\textbf{Multi-News}}
         & \multicolumn{2}{c}{\textbf{IELTS}} \\
        \cmidrule(lr){3-4}\cmidrule(lr){5-6}\cmidrule(lr){7-8}
         &
         & \textbf{AUC} & \textbf{R-L}
         & \textbf{AUC} & \textbf{R-L}
         & \textbf{AUC} & \textbf{Jdg.} \\
        \midrule
        
        \multicolumn{8}{c}{\textbf{Llama-3}} \\
        \midrule
        
        \multirow{4}{*}{\shortstack[l]{Fake-\\SpotAI}}
         & \graycell{Vanilla}
         & \graycell{\best{100.0}} & \graycell{22.9}
         & \graycell{\best{100.0}} & \graycell{26.5}
         & \graycell{99.7}  & \graycell{6.77} \\
         & ELI5 \puppeticon
         & \best{100.0} & \best{25.2}\sigup & \best{100.0} & \best{27.1}\sigup & 99.8 & 6.74 \\
         & MN \puppeticon
         & \best{100.0} & 23.9\sigup & \best{100.0} & 26.1 & \best{99.9} & \best{7.32} \\
         & IELTS \puppeticon
         & \best{100.0} & 23.7\sigup & \best{100.0} & 26.3 & \best{99.9} & \best{7.32}\sigup \\
        \midrule
        
        \multirow{4}{*}{MAGE}
         & \graycell{Vanilla}
         & \graycell{\best{100.0}} & \graycell{22.9}
         & \graycell{97.2}  & \graycell{26.5}
         & \graycell{89.4}  & \graycell{6.77} \\
         & ELI5 \puppeticon
         & \best{100.0} & \best{24.5}\sigup & \best{97.3} & 27.2\sigup & 90.0 & 6.66 \\
         & MN \puppeticon
         & \best{100.0} & 23.5\sigup & 95.4 & \best{29.2}\sigup & 90.1 & 6.68 \\
         & IELTS \puppeticon
         & \best{100.0} & 22.1\sigdown & 96.7 & 26.0 & \best{94.2} & \best{7.31}\sigup \\
        \midrule
        
        \multicolumn{8}{c}{\textbf{Qwen3}} \\
        \midrule
        
        \multirow{4}{*}{\shortstack[l]{Fake-\\SpotAI}}
         & \graycell{Vanilla}
         & \graycell{\best{100.0}} & \graycell{24.5}
         & \graycell{\best{100.0}} & \graycell{24.0}
         & \graycell{98.9}  & \graycell{7.63} \\
         & ELI5 \puppeticon
         & \best{100.0} & \best{25.7}\sigup & \best{100.0} & \best{24.4}\sigup & 99.4 & \best{7.67} \\
         & MN \puppeticon
         & \best{100.0} & 25.0\sigup & \best{100.0} & 24.1 & 99.4 & 7.64 \\
         & IELTS \puppeticon
         & \best{100.0} & 24.9\sigup & \best{100.0} & 24.0 & \best{99.6} & 7.63 \\
        \midrule
        
        \multirow{4}{*}{MAGE}
         & \graycell{Vanilla}
         & \graycell{\best{100.0}} & \graycell{24.5}
         & \graycell{97.5}  & \graycell{24.0}
         & \graycell{91.3}  & \graycell{7.63} \\
         & ELI5 \puppeticon
         & \best{100.0} & 25.3\sigup & 97.2 & 24.4 & 91.2 & 7.65 \\
         & MN \puppeticon
         & \best{100.0} & 24.4 & 97.4 & \best{25.9}\sigup & 91.2 & 7.63 \\
         & IELTS \puppeticon
         & \best{100.0} & \best{25.5}\sigup & \best{98.1} & 24.6\sigup & \best{95.3} & \best{7.85}\sigup \\
        
        \bottomrule
        
        \CodeAfter
        \begin{tikzpicture}[overlay]
          \def\xpad{2.0pt}
          \def\ypadtop{1.6pt}
          \def\ypadbot{0.8pt}
          \def\ynotch{1.2pt}
        
          \draw[line width=0.4pt, rounded corners=2pt]
            ([xshift=-\xpad,yshift=\ypadtop]5-5.north west) --
            ([xshift=\xpad,yshift=\ypadtop]5-8.north east) --
            ([xshift=\xpad,yshift=-\ypadbot]6-8.south east) --
            ([xshift=-\xpad,yshift=-\ypadbot]6-7.south west) --
            ([xshift=-\xpad,yshift=\ypadtop]6-7.north west) --
            ([xshift=-\xpad,yshift=-\ypadbot]5-5.south west) -- cycle;
        
          \draw[line width=0.4pt, rounded corners=2pt]
            ([xshift=-\xpad,yshift=\ypadtop]6-3.north west) --
            ([xshift=\xpad,yshift=\ypadtop]6-4.north east) --
            ([xshift=\xpad,yshift=\ynotch]7-4.north east) --
            ([xshift=\xpad,yshift=\ynotch]7-6.north east) --
            ([xshift=\xpad,yshift=-\ypadbot]7-6.south east) --
            ([xshift=-\xpad,yshift=-\ypadbot]7-3.south west) --
            ([xshift=-\xpad,yshift=-\ypadbot]6-3.south west) -- cycle;
        
          \draw[line width=0.4pt, rounded corners=2pt]
            ([xshift=-\xpad,yshift=\ypadtop]9-5.north west) --
            ([xshift=\xpad,yshift=\ypadtop]9-8.north east) --
            ([xshift=\xpad,yshift=-\ypadbot]10-8.south east) --
            ([xshift=-\xpad,yshift=-\ypadbot]10-7.south west) --
            ([xshift=-\xpad,yshift=\ypadtop]10-7.north west) --
            ([xshift=-\xpad,yshift=-\ypadbot]9-5.south west) -- cycle;
        
          \draw[line width=0.4pt, rounded corners=2pt]
            ([xshift=-\xpad,yshift=\ypadtop]10-3.north west) --
            ([xshift=\xpad,yshift=\ypadtop]10-4.north east) --
            ([xshift=\xpad,yshift=\ynotch]11-4.north east) --
            ([xshift=\xpad,yshift=\ynotch]11-6.north east) --
            ([xshift=\xpad,yshift=-\ypadbot]11-6.south east) --
            ([xshift=-\xpad,yshift=-\ypadbot]11-3.south west) --
            ([xshift=-\xpad,yshift=-\ypadbot]10-3.south west) -- cycle;
        
          \draw[line width=0.4pt, rounded corners=2pt]
            ([xshift=-\xpad,yshift=\ypadtop]14-5.north west) --
            ([xshift=\xpad,yshift=\ypadtop]14-8.north east) --
            ([xshift=\xpad,yshift=-\ypadbot]15-8.south east) --
            ([xshift=-\xpad,yshift=-\ypadbot]15-7.south west) --
            ([xshift=-\xpad,yshift=\ypadtop]15-7.north west) --
            ([xshift=-\xpad,yshift=-\ypadbot]14-5.south west) -- cycle;
        
          \draw[line width=0.4pt, rounded corners=2pt]
            ([xshift=-\xpad,yshift=\ypadtop]15-3.north west) --
            ([xshift=\xpad,yshift=\ypadtop]15-4.north east) --
            ([xshift=\xpad,yshift=\ynotch]16-4.north east) --
            ([xshift=\xpad,yshift=\ynotch]16-6.north east) --
            ([xshift=\xpad,yshift=-\ypadbot]16-6.south east) --
            ([xshift=-\xpad,yshift=-\ypadbot]16-3.south west) --
            ([xshift=-\xpad,yshift=-\ypadbot]15-3.south west) -- cycle;
        
          \draw[line width=0.4pt, rounded corners=2pt]
            ([xshift=-\xpad,yshift=\ypadtop]18-5.north west) --
            ([xshift=\xpad,yshift=\ypadtop]18-8.north east) --
            ([xshift=\xpad,yshift=-\ypadbot]19-8.south east) --
            ([xshift=-\xpad,yshift=-\ypadbot]19-7.south west) --
            ([xshift=-\xpad,yshift=\ypadtop]19-7.north west) --
            ([xshift=-\xpad,yshift=-\ypadbot]18-5.south west) -- cycle;
        
          \draw[line width=0.4pt, rounded corners=2pt]
            ([xshift=-\xpad,yshift=\ypadtop]19-3.north west) --
            ([xshift=\xpad,yshift=\ypadtop]19-4.north east) --
            ([xshift=\xpad,yshift=\ynotch]20-4.north east) --
            ([xshift=\xpad,yshift=\ynotch]20-6.north east) --
            ([xshift=\xpad,yshift=-\ypadbot]20-6.south east) --
            ([xshift=-\xpad,yshift=-\ypadbot]20-3.south west) --
            ([xshift=-\xpad,yshift=-\ypadbot]19-3.south west) -- cycle;
        
        \end{tikzpicture}
    \end{NiceTabular}
    \caption{
    Detection and task performance across alternative detectors, for both Llama-3 and Qwen3.
    Rows: task used for fine-tuning; columns: evaluation benchmarks.
    Boxes highlight OOD evaluations (off-diagonal).
    \sigupcap{} and \sigdowncap{} denote significant improvement and degeneration on task performance (paired $t$-test, $\alpha=0.05$), respectively.
    }
    \label{tab:multi_detector_ood}
\end{table}

\begin{table}[t]
    \setlength{\tabcolsep}{1mm}
    \centering
    \small
    \begin{NiceTabular}{lcccccc}
        \toprule
         & \multicolumn{2}{c}{\textbf{ELI5}} & \multicolumn{2}{c}{\textbf{Multi-News}} & \multicolumn{2}{c}{\textbf{IELTS}} \\
        \cmidrule(lr){2-3}\cmidrule(lr){4-5}\cmidrule(lr){6-7}
        \textbf{Method} & \textbf{AUR} & \textbf{R-L} & \textbf{AUR} & \textbf{R-L} & \textbf{AUR} & \textbf{Jdg.} \\
        \midrule

        \multicolumn{7}{c}{\textbf{Llama-3}}\\
        \midrule
        \multicolumn{7}{c}{\textsc{Watermarking baselines}}\\
        \midrule

        KGW & 70.1 & 23.9\sigup & 71.6 & 23.2\sigdown & 72.0 & 5.45\sigdown \\
        SynthID & 55.7 & 23.6 & 54.3 & 23.0\sigdown & 56.5 & 5.41\sigdown \\
        Unigram & 32.3 & 23.3 & 61.3 & 23.1\sigdown & 45.3 & 5.26\sigdown \\
        EXPGumbel & 65.8 & 22.4 & 54.2 & 21.9\sigdown & 49.4 & 4.84 \\
        K-SemStamp & 60.6 & 23.6\sigup & 65.1 & 22.9\sigdown & 54.7 & 5.54\sigdown \\

        \midrule
        \multicolumn{7}{c}{\textsc{Ours}}\\
        \midrule

        \rowcolor{lightgrayrow}
        Vanilla & 98.8 & 23.3\sigup & 98.8 & 23.3\sigdown & 92.9 & 5.46\sigdown \\
        ELI5\puppeticon & \best{100.0} & \best{24.9}\sigdown & 98.9 & 23.9\sigdown & 95.0 & 5.62\sigdown \\
        Multi-News\puppeticon & 99.4 & 24.2\sigdown & \best{99.4} & \best{24.5}\sigdown & \best{97.2} & 5.56\sigdown \\
        IELTS\puppeticon & 99.8 & 23.7\sigdown & 98.8 & 23.6\sigdown & 96.2 & \best{5.97}\sigdown \\

        \addlinespace[4pt]

        \midrule
        \multicolumn{7}{c}{\textbf{Qwen3}}\\
        \midrule
        \multicolumn{7}{c}{\textsc{Watermarking baselines}}\\
        \midrule

        KGW & 59.2 & 24.4 & 60.9 & 22.9\sigdown & 66.1 & 5.94\sigdown \\
        SynthID & 55.0 & 24.4 & 61.2 & 22.9\sigdown & 50.8 & 5.91\sigdown \\
        Unigram & 59.8 & 24.2 & 80.0 & 23.1\sigdown & 80.2 & 5.80\sigdown \\
        EXPGumbel & 60.1 & 23.2\sigdown & 56.9 & 20.9\sigdown & 58.2 & 5.02\sigup \\
        K-SemStamp & 60.7 & 24.4 & 65.6 & 22.5\sigdown & 60.7 & 5.78\sigdown \\

        \midrule
        \multicolumn{7}{c}{\textsc{Ours}}\\
        \midrule

        \rowcolor{lightgrayrow}
        Vanilla & 98.6 & 24.4 & 98.4 & 23.1\sigdown & 90.3 & 5.85\sigdown \\
        ELI5\puppeticon & \best{100.0} & \best{24.9}\sigdown & 98.4 & 23.1\sigdown & 92.5 & 5.95\sigdown \\
        Multi-News\puppeticon & 99.4 & 24.2\sigdown & 98.4 & \best{23.8}\sigdown & 94.3 & 5.83\sigdown \\
        IELTS\puppeticon & 99.7 & \best{24.9}\sigdown & \best{98.7} & 23.2\sigdown & \best{96.5} & \best{6.10}\sigdown \\

        \bottomrule

        \CodeAfter
        \begin{tikzpicture}[overlay]
          \def\xpad{2.0pt}
          \def\ypadtop{1.6pt}
          \def\ypadbot{0.8pt}
          \def\ynotch{1.2pt}

          \draw[line width=0.4pt, rounded corners=2pt]
            ([xshift=-\xpad,yshift=\ypadtop]12-4.north west) --
            ([xshift=\xpad,yshift=\ypadtop]12-7.north east) --
            ([xshift=\xpad,yshift=-\ypadbot]13-7.south east) --
            ([xshift=-\xpad,yshift=-\ypadbot]13-6.south west) --
            ([xshift=-\xpad,yshift=\ypadtop]13-6.north west) --
            ([xshift=-\xpad,yshift=-\ypadbot]12-4.south west) -- cycle;

          \draw[line width=0.4pt, rounded corners=2pt]
            ([xshift=-\xpad,yshift=\ypadtop]13-2.north west) --
            ([xshift=\xpad,yshift=\ypadtop]13-3.north east) --
            ([xshift=\xpad,yshift=\ynotch]14-3.north east) --
            ([xshift=\xpad,yshift=\ynotch]14-5.north east) --
            ([xshift=\xpad,yshift=-\ypadbot]14-5.south east) --
            ([xshift=-\xpad,yshift=-\ypadbot]14-2.south west) --
            ([xshift=-\xpad,yshift=-\ypadbot]13-2.south west) -- cycle;

          \draw[line width=0.4pt, rounded corners=2pt]
            ([xshift=-\xpad,yshift=\ypadtop]24-4.north west) --
            ([xshift=\xpad,yshift=\ypadtop]24-7.north east) --
            ([xshift=\xpad,yshift=-\ypadbot]25-7.south east) --
            ([xshift=-\xpad,yshift=-\ypadbot]25-6.south west) --
            ([xshift=-\xpad,yshift=\ypadtop]25-6.north west) --
            ([xshift=-\xpad,yshift=-\ypadbot]24-4.south west) -- cycle;

          \draw[line width=0.4pt, rounded corners=2pt]
            ([xshift=-\xpad,yshift=\ypadtop]25-2.north west) --
            ([xshift=\xpad,yshift=\ypadtop]25-3.north east) --
            ([xshift=\xpad,yshift=\ynotch]26-3.north east) --
            ([xshift=\xpad,yshift=\ynotch]26-5.north east) --
            ([xshift=\xpad,yshift=-\ypadbot]26-5.south east) --
            ([xshift=-\xpad,yshift=-\ypadbot]26-2.south west) --
            ([xshift=-\xpad,yshift=-\ypadbot]25-2.south west) -- cycle;

        \end{tikzpicture}
    \end{NiceTabular}
    \caption{
        Detection and task performance results after the Dipper paraphrasing attack, for Llama-3 and Qwen3.
        Rows: task used for fine-tuning; columns: evaluation benchmarks.
        Bold: best per column.
        Boxes highlight OOD evaluations (off-diagonal).
        \sigupcap{} and \sigdowncap{} denote significant improvement and degeneration on task performance relative to before the attack (paired $t$-test, $\alpha=0.05$), respectively.
    }
    \label{tab:paraphrase_full}
\end{table}

\section{Full Paraphrasing Attack Results} \label{app:paraphrase}

\begin{table}[t]
    \setlength{\tabcolsep}{1mm}
    \centering
    \small
    \begin{tabular}{llcccccc}
        \toprule
         & & \multicolumn{2}{c}{\textbf{ELI5}}
         & \multicolumn{2}{c}{\textbf{Multi-News}}
         & \multicolumn{2}{c}{\textbf{IELTS}} \\
        \cmidrule(lr){3-4}\cmidrule(lr){5-6}\cmidrule(lr){7-8}
        \textbf{Detector} & \textbf{Method}
         & \textbf{AUC} & \textbf{R-L}
         & \textbf{AUC} & \textbf{R-L}
         & \textbf{AUC} & \textbf{Jdg.} \\
        \midrule
        
        \multicolumn{8}{c}{\textbf{Llama-3}}\\
        \midrule
        
        \multirow{2}{*}{\shortstack[l]{Fake-\\SpotAI}}
         & \graycell{Vanilla}
         & \graycell{\best{100.0}} & \graycell{23.3}
         & \graycell{97.8}  & \graycell{\best{23.3}}
         & \graycell{95.9}   & \graycell{5.46} \\
         & PUPPET\puppeticon
         & \best{100.0} & \best{24.9}
         & \best{99.9}  & 23.2\sigdown
         & \best{96.1}   & \best{5.93}\sigdown \\
        \midrule

        \multirow{2}{*}{MAGE}
         & \graycell{Vanilla}
         & \graycell{\best{99.9}} & \graycell{23.3}
         & \graycell{\best{86.9}}  & \graycell{23.3}
         & \graycell{94.0}  & \graycell{\best{5.46}} \\
         & PUPPET\puppeticon
         & 99.7 & \best{24.3}\sigdown
         &  84.9 & \best{25.2}\sigdown
         &  \best{95.1} & 5.34\sigdown \\
        
        \addlinespace[4pt]
        
        \midrule
        \multicolumn{8}{c}{\textbf{Qwen3}}\\
        \midrule
        
        \multirow{2}{*}{\shortstack[l]{Fake-\\SpotAI}}
         & \graycell{Vanilla}
         & \graycell{\best{100.0}} & \graycell{24.4}
         & \graycell{98.9}  & \graycell{\best{23.1}}
         & \graycell{\best{94.5}}   & \graycell{5.85} \\
         & PUPPET\puppeticon
         & \best{100.0} & \best{25.0}\sigdown
         & \best{99.6}  & 22.7\sigdown
         &  94.4  & \best{6.02}\sigdown \\
        \midrule

        \multirow{2}{*}{MAGE}
         & \graycell{Vanilla}
         & \graycell{\best{99.9}} & \graycell{24.4}
         & \graycell{\best{89.1}}  & \graycell{\best{23.5}}
         & \graycell{94.5}  & \graycell{\best{5.85}} \\
         & PUPPET\puppeticon
         & 99.7 & \best{24.8}\sigdown
         &  88.3 & \best{23.5}\sigdown
         &  \best{94.6} & 5.49\sigdown \\
        
        \bottomrule
    \end{tabular}
    \caption{
        Detection and task performance after the Dipper paraphrasing attack,
        across FakeSpotAI and MAGE detectors, for Llama-3 and Qwen3.
        Rows: task used for fine-tuning; columns: evaluation benchmarks.
        \sigupcap{} and \sigdowncap{} denote significant improvement and degeneration on task performance relative to before the attack (paired $t$-test, $\alpha=0.05$), respectively.
    }
    \label{tab:paraphrase_multi_detector}
\end{table}

This appendix presents full paraphrasing attack results complementing the summary in \S~\ref{sec:robustness}.
\S~\ref{app:paraphrase_openai} covers results when PUPPET is fine-tuned with the OpenAI Detector, and \S~\ref{app:paraphrase_multi} when it is fine-tuned with FakeSpotAI and MAGE.

\subsection{Against the OpenAI Detector} \label{app:paraphrase_openai}
Table~\ref{tab:paraphrase_full} reports detection and task performance after applying Dipper as a paraphrasing attack, across both models and all benchmarks.
Pre-paraphrasing results are reported in Table~\ref{tab:result_main} and Table~\ref{tab:robustness_ood} for Llama-3, and in Table~\ref{tab:result_main_qwen_indomain} and Table~\ref{tab:robustness_ood_qwen} for Qwen3.
AUROC drops sharply after paraphrasing for all watermarking baselines (e.g., KGW on ELI5 with Llama-3: $99.6 \to 70.1$), as Dipper's surface-level token substitutions destroy the statistical patterns embedded by watermarking.
In contrast, PUPPET's AUROC is largely unaffected, consistently outperforming the Vanilla baseline in both detection and task performance---with ID results again roughly exceeding OOD ones, across both Llama-3 and Qwen3.
This pattern suggests that PUPPET captures detection-relevant features that are not easily erased by surface-level rewriting.

Task performance declines after paraphrasing across all methods.
This degradation reflects a property of Dipper itself: as a T5-based model, Dipper produces text of lower quality than Llama-3 or Qwen3, causing semantic degradation regardless of the original generation method.
Nevertheless, in the ID setting PUPPET still retains higher task performance than Vanilla across all configurations.

\subsection{Against FakeSpotAI and MAGE} \label{app:paraphrase_multi}
Table~\ref{tab:paraphrase_multi_detector} reports ID detection and task performance after the Dipper paraphrasing attack for PUPPET fine-tuned with FakeSpotAI and MAGE.
Pre-paraphrasing results are reported in Table~\ref{tab:robustness_detector} for Llama-3, and in Table~\ref{tab:multi_detector_ood} for Qwen3.

Unlike the consistent pattern observed with the OpenAI Detector, AUROC changes after paraphrasing vary across detectors.
For FakeSpotAI, the largest degradation occurs on Qwen3's IELTS, dropping from $99.6$ to $94.4$ (${-}5.2$); for MAGE, it is even larger, with Llama-3's Multi-News falling from $95.4$ to $84.9$ (${-}10.5$).
This variability likely reflects whether the text produced by Dipper retains features that each detector relies on:
when a detector's signal depends on surface-level patterns easily destroyed by rewriting, detection collapses;
when those features are higher-level or incidentally preserved, detection holds.
Task performance, in contrast, degrades almost uniformly and significantly across detectors, though PUPPET still matches or exceeds Vanilla in $8$ of the $12$ settings.
These observations suggest that PUPPET learns detection-relevant features that are not merely surface-level and thus survive rewriting, while also showing that post-paraphrasing detection depends on each detector's sensitivity to such perturbations.

\begin{figure*}[t]
    \centering
    \includegraphics[width=0.9\linewidth]{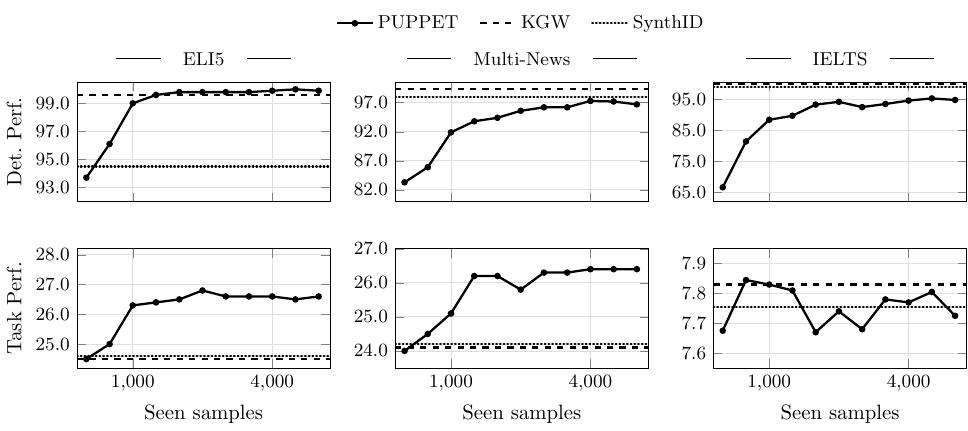}
    \caption{
        Detection performance (Det.\ Perf., AUROC; top row) and task performance
        (Task Perf., ROUGE-L for ELI5 and Multi-News, judge score for IELTS; bottom row)
        as a function of training samples seen for Qwen3 across all three benchmarks.
        (500 samples per dot)
        Dashed and dotted lines indicate KGW and SynthID baselines, respectively.
    }
    \label{fig:seen_samples_curve_qwen3}
\end{figure*}

\section{Training Efficiency: Qwen3-8B} \label{app:practicality_of_training}

Figure~\ref{fig:seen_samples_curve_qwen3} shows sample-efficiency curves for Qwen3, complementing the Llama-3 results in Figure~\ref{fig:seen_samples_curve_llama3}.
Qwen3 follows a pattern similar to Llama-3: both detection and task performance converge within a few thousand DPO training examples on ELI5 and Multi-News.
On IELTS, where the Vanilla detection performance is considerably lower ($66.6$, vs.\ $93.7$ for ELI5 and $83.3$ for Multi-News), requiring more samples to approach its ceiling.
The task performance curve for Qwen3 on IELTS is notably unstable and fails to converge to a clear plateau within the observed range, which we attribute to the OpenAI Detector's difficulty detecting in this setting.

\section{Reward Composition Ablation: Qwen3-8B} \label{app:reward_ablation}

\begin{figure}[t]
    \centering
    \includegraphics[width=\linewidth]{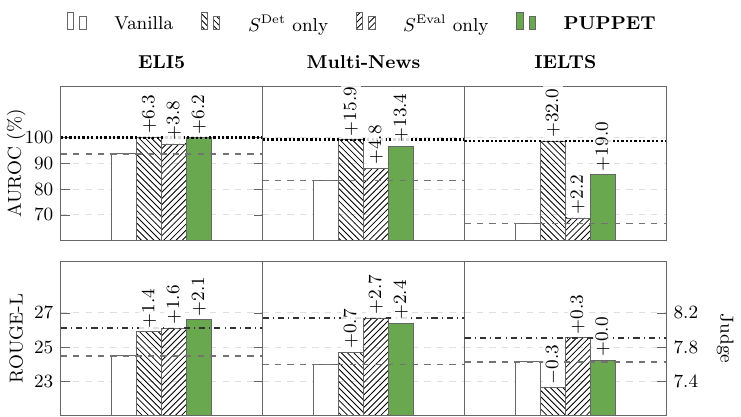}
    \caption{
        Reward ablation (Qwen3).
        Bar labels: point-wise changes from Vanilla.
        In the AUROC panels, dotted lines mark the $S^{\text{Det}}$-only ceiling;
        in the task panels, dash-dotted lines mark the $S^{\text{Eval}}$-only ceiling;
        dashed lines mark the Vanilla baseline in all panels.
        ROUGE-L (left) and Judge (right) task axes are proportionally scaled.
    }
    \label{fig:reward_ablation_qwen}
\end{figure}

\begin{table}[t]
    \centering
    \small
    \setlength{\tabcolsep}{1mm}
    \begin{tabular}{llccc}
    \toprule
    \textbf{Target} & \textbf{Contrast}
      & \textbf{ELI5} & \textbf{Multi-News} & \textbf{IELTS} \\
    \midrule
    \rowcolor{lightgrayrow}
    Qwen3   & Llama-3 & 41.01 {\scriptsize($-$0.23)} & 27.74 {\scriptsize($-$0.82)} & 22.29 {\scriptsize($-$0.96)} \\
    Qwen3\puppeticon    & Llama-3 & \best{87.98} {\scriptsize(1.22)} & 59.72 {\scriptsize(0.31)} & \best{71.09} {\scriptsize(0.78)} \\
    \bottomrule
    \end{tabular}
    \caption{
      Model-attribution AUROC (OpenAI Detector), treating \textit{Target} as the positive class and \textit{Contrast} as the negative.
      Parentheses: Cohen's $d$; models are trained per task.
    }
    \label{tab:distinguish_models_full}
\end{table}

Figure~\ref{fig:reward_ablation_qwen} reports reward composition ablation results for Qwen3, complementing the Llama-3 results in Figure~\ref{fig:reward_ablation}.

The overall pattern is consistent with Llama-3: $S^{\text{Det}}$ only achieves large AUROC gains at the cost of task performance, $S^{\text{Eval}}$ only improves task without lifting detection, and PUPPET improves on both axes.
On ELI5 and Multi-News, PUPPET recovers---and in some cases exceeds---the single-objective gains: $84.3$--$98.4\%$ of the $S^{\text{Det}}$-only detection gain and $88.9$--$131.3\%$ of the $S^{\text{Eval}}$-only task gain.
A notable difference from Llama-3 appears on IELTS, where PUPPET's detection gain recovery drops to $59.4\%$ (vs.\ $95.3\%$ for Llama-3).
This is likely because the Vanilla AUROC is more than 20 points lower than Llama-3's ($66.6$ vs.\ $87.0$ for Llama-3), so a fixed $\alpha = 0.5$ was insufficient to fully optimize detection performance under this setting.

\begin{figure*}[t]
    \centering
    \includegraphics[width=0.87\linewidth]{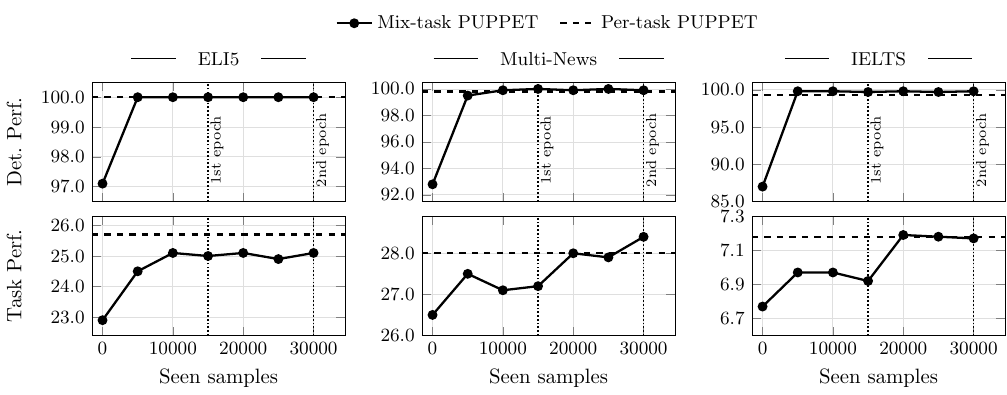}
    \caption{
        Multi-task learning: Detection performance (Det.\ Perf., AUROC; top row) and task performance
        (Task Perf., ROUGE-L for ELI5 and Multi-News, judge score for IELTS; bottom row)
        as a function of training samples seen for Llama-3 across all three benchmarks
        (5,000 samples per dot).
        Dashed lines indicate the standard, per-task PUPPET.
    }
    \label{fig:mix_dataset}
\end{figure*}

\begin{figure}[t]
    \centering
    \includegraphics[width=\linewidth]{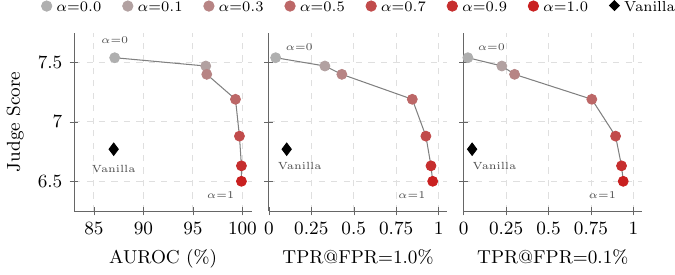}
    \caption{
        $\alpha$ ablation on IELTS for Llama-3.
        $\alpha{=}0$ corresponds to $S^{\text{Eval}}$ only; $\alpha{=}1$ corresponds to $S^{\text{Det}}$ only.
        As $\alpha$ increases, the reward shifts toward favoring detection.
    }
    \label{fig:alpha_ablation_ielts}
\end{figure}

\section{Model-Attribution Results for Qwen3} \label{app:model_attribution}
Table~\ref{tab:distinguish_models_full} reports model-attribution AUROC from the Qwen3 perspective, complementing the Llama-3 results in Table~\ref{tab:distinguish_models}.

At the vanilla baseline, the detector achieves AUROC of $41.01$ on ELI5 (below chance from this perspective) and $27.74$--$22.29$ on Multi-News and IELTS---the mirror image of the Llama-3 baseline, confirming that the asymmetry is an artifact of the uncontrolled detector rather than a meaningful signal in either direction.
Applying PUPPET to Qwen3 substantially raises AUROC against Vanilla Llama-3 on ELI5 ($87.98$) and IELTS ($71.09$), though the gain on Multi-News ($59.72$) is more modest.
These results are consistently lower than the PUPPET Llama-3 versus Vanilla Qwen3 counterparts ($94.47$, $96.94$, $99.14$; Table~\ref{tab:distinguish_models}).
We attribute this asymmetry to the baseline machine-class likelihood distributions: Vanilla Llama-3 already occupies a high machine-class likelihood region,
whereas Vanilla Qwen3 lies in a lower one (Table~\ref{tab:result_main} for Llama-3 and Table~\ref{tab:result_main_qwen_indomain} for Qwen3).
When PUPPET shifts Qwen3's distribution upward, it reaches a level comparable to Vanilla Llama-3 rather than surpassing it, leaving the two distributions insufficiently separated---most strikingly on Multi-News, where $d = 0.31$, versus $d = 3.18$ for the reverse comparison (PUPPET Llama-3 vs.\ Vanilla Qwen3), which explains the latter's markedly higher AUROC in Table~\ref{tab:distinguish_models}.
Overall, these results suggest that PUPPET can also support post-release model tracking by model providers when applied to Qwen3.

\section{Finer-Grained $\alpha$ Ablation on IELTS} \label{app:alpha_ablation_ielts}

Across base LLMs and detectors, Vanilla detection performance on IELTS is consistently lower than on the other benchmarks,
which we observe makes joint optimization more challenging (Table~\ref{tab:result_main}, Table~\ref{tab:robustness_detector}, and Table~\ref{tab:tpr_at_fpr} for Llama-3).
We hypothesized that weighting the detection reward might reveal a better operating point, so we extended the $\alpha \in \{0.0, 0.5, 1.0\}$ ablation reported for Llama-3 in Figure~\ref{fig:reward_ablation} with additional runs at $\alpha \in \{0.1, 0.3, 0.7, 0.9\}$.
The results are shown in Figure~\ref{fig:alpha_ablation_ielts}.
Across all detection metrics, we observe a trade-off against Judge score.
For AUROC, this trade-off is relatively mild once $\alpha \geq 0.1$, whereas for TPR it is more pronounced.
In summary, PUPPET clearly occupies a more favorable region of it than Vanilla.

\section{Multi-Task Learning} \label{app:mix_dataset}

Throughout this paper, PUPPET is fine-tuned separately on each benchmark.
In practice, however, a single model is often expected to handle multiple tasks.
While our OOD evaluations already address cross-task generalization, they do not involve directly training on multiple tasks at once.
We therefore fine-tune Llama-3 on a shuffled mixture of all three task datasets ($5,000\times3=15,000$ samples in total) for 2 epochs,
with all other hyperparameters identical to those in \S~\ref{app:details}.

Results are shown in Figure~\ref{fig:mix_dataset}.
Detection performance reaches a level comparable to per-task PUPPET after only $5,000$ samples, equal in size to just one of the three datasets.
Task performance, by contrast, is harder to optimize at once:
Multi-News and IELTS reach the per-task level only around the $20,000$ samples, $5,000$ samples beyond $1$ epoch,
and ELI5 still falls short by about $1$ point even after the full $2$ epochs.
We attribute this asymmetry to the nature of the supervision signals: detection performance is supervised by the OpenAI Detector's machine-class likelihood shared across all tasks,
whereas task performance relies on heterogeneous, per-task metrics (ROUGE-L for ELI5/Multi-News vs.\ LLM-as-a-Judge for IELTS) spanning different domains, whose conflicting optimization directions are known to hinder joint multi-task optimization~\citep{gradientSurgeryForMultiTaskLearning, selfImprovementTowardsParetoOptimality}.

\end{document}